\documentclass{article}

 \PassOptionsToPackage{numbers, compress}{natbib}

\usepackage{arxiv}

\usepackage{amsmath,amsfonts,bm, amsthm,algorithm, algpseudocode}

\newcommand{\id}{\mathbf{Id}}

\def\Figref#1{Fig.~\ref{#1}}

\def\Secref#1{Sec.~\ref{#1}}

\def\eqref#1{eq.~\ref{#1}}
\def\Eqref#1{Eq.~\ref{#1}}

\def\1{\bm{1}}

\def\rva{{\mathbf{a}}}
\def\rvb{{\mathbf{b}}}
\def\rvc{{\mathbf{c}}}

\def\rvu{{\mathbf{i}}}

\def\rvm{{\mathbf{m}}}

\def\rvp{{\mathbf{p}}}

\def\rvu{{\mathbf{u}}}
\def\rvv{{\mathbf{v}}}

\def\rvx{{\mathbf{x}}}

\def\rvz{{\mathbf{z}}}

\def\rmA{{\mathbf{A}}}
\def\rmB{{\mathbf{B}}}
\def\rmC{{\mathbf{C}}}

\def\rmU{{\mathbf{U}}}
\def\rmV{{\mathbf{V}}}

\def\mM{{\bm{M}}}

\DeclareMathAlphabet{\mathsfit}{\encodingdefault}{\sfdefault}{m}{sl}
\SetMathAlphabet{\mathsfit}{bold}{\encodingdefault}{\sfdefault}{bx}{n}

\def\gD{{\mathcal{D}}}

\def\gP{{\mathcal{P}}}

\def\sP{{\mathbb{P}}}

\def\sR{{\mathbb{R}}}

\newcommand{\E}{\mathbb{E}}

\newcommand{\R}{\mathbb{R}}

\theoremstyle{plain}

\theoremstyle{remark}

\theoremstyle{definition}

\theoremstyle{plain}

\theoremstyle{plain}

\theoremstyle{definition}

\providecommand{\corollaryname}{Corollary}
\providecommand{\lemmaname}{Lemma}
\providecommand{\problemname}{Problem}
\providecommand{\remarkname}{Remark}
\providecommand{\theoremname}{Theorem}

\usepackage[utf8]{inputenc} 
\usepackage[T1]{fontenc}    
\usepackage{booktabs}       
\usepackage{amsfonts}       
\usepackage{nicefrac}       
\usepackage{microtype}      
\usepackage{enumitem}
\usepackage{graphicx}
\usepackage{mathtools}
\usepackage{gensymb}
\usepackage{amssymb}
\usepackage{color}
\usepackage{hyperref}       
\usepackage{url}            
\usepackage{multirow}
\usepackage{todonotes}
\usepackage{booktabs}
\usepackage{threeparttable}
\usepackage{wrapfig}
\usepackage{array}
\usepackage{pgfplots}
\usepackage{subcaption}
\usepackage{booktabs}
\usepackage{blindtext}
\usepackage[export]{adjustbox}
\usepackage{rotating}
\usepackage{tcolorbox}
\usepackage{epstopdf}

\newcommand{\wa}{0.07}
\newcommand{\etal}{et al.}
\definecolor{amber}{rgb}{1.0, 0.75, 0.0}
\definecolor{ube}{rgb}{0.53, 0.47, 0.76}
\definecolor{seabornback}{HTML}{EAEAF2}

\newtheorem{theorem}{Theorem}[section]

\title{Encoding Invariances in Deep Generative Models}

\author{
Viraj Shah\thanks{Equal contribution}\hspace{4pt}\thanks{Dept. of Electrical and Computer Engineering} \qquad Ameya Joshi\footnotemark[1]\hspace{4pt}\footnotemark[2] \qquad Sambuddha Ghosal\thanks{Dept. of Mechanical Engineering} \qquad Balaji Pokuri\footnotemark[3]\\
Iowa State University\\
\texttt{\{viraj, ameya, sghosal, balajip\}@iastate.edu}
\And
Soumik Sarkar \footnotemark[3] \qquad Baskar Ganapathysubramanian \footnotemark[3] \qquad Chinmay Hegde \footnotemark[2]\\
Iowa State University, Ames \\
\texttt{\{soumiks, baskarg, chinmay\}@iastate.edu}
}

\begin{document}

\maketitle

\begin{abstract}
    Reliable training of generative adversarial networks (GANs)  typically require massive datasets in order to model complicated distributions. However, in several applications, training samples obey invariances that are \textit{a priori} known; for example, in complex physics simulations, the training data obey universal laws encoded as well-defined mathematical equations. In this paper, we propose a new generative modeling approach, InvNet, that can efficiently model data spaces with known invariances. We devise an adversarial training algorithm to encode them into data distribution. We validate our framework in three experimental settings: generating images with fixed motifs; solving nonlinear partial differential equations (PDEs); and reconstructing two-phase microstructures with desired statistical properties. We complement our experiments with several theoretical results.
\end{abstract}

\section{Introduction}
 
\paragraph{Motivation.} Generative Adversarial Networks (GANs) have proven to be highly successful in synthesizing samples arising from complex distributions, including
face images~\cite{Radford2016UnsupervisedRL}, content generation~\cite{Jin2017TowardsTA}, image translation~\cite{Isola2017ImagetoImageTW}, style transfer~\cite{CycleGAN2017}, and many others. However, a well-known issue of GANs is that they incur dramatically \emph{high sample complexity}; for example, the recent work of \cite{Brock2019LargeSG} requires 14 million training images and is trained over $\sim$24K TPU-hours, which is beyond the reach of normal computing environments. 
 
This challenge partly arises due to the purely unsupervised nature of GAN training. Intuitively, GAN models start learning from scratch, and require lots of training examples before they learn to reproduce essential features (or {\it invariances}) present in the training data.  However, in several applications, invariances that the generated samples should exhibit are explicitly known prior to training. 
For example, in scientific simulations, the data samples often obey (universal) laws encoded in the form of well-defined mathematical equations, or obey other statistical or geometric constraints. How, then, should GAN models best leverage such prior knowledge of invariance information? 

\paragraph{Our contributions.} In this paper, we propose a systematic extension of GANs that can synthesize data samples obeying pre-specified, analytically defined invariances. We call our new generative models \emph{Invariance Networks}, or InvNets. Our approach subsumes the standard GAN setting; InvNets can learn \emph{implicit} invariances from the training data (similar to GANs), but also enforce \emph{explicit} invariances. 
Similar to GANs, we pose the InvNet training problem as a minimax game. Training an InvNet requires some care, since interesting new challenges arise. We propose a three-way alternating-optimization style training algorithm that gives useful, stable results across a wide range of invariances and training data. We illustrate InvNet performance in three diverse test applications:

\textbf{1. \emph{Encoding motifs in images.}} As a stylized application, we design an InvNet model for stitching a special, pre-defined pattern (or motif) onto all synthesized image samples. While existing (non-GAN) methods such as image blending can achieve this with ease, this toy application serves both as a proof-of-concept, and reveals interesting theoretical aspects of the training dynamics.

\begin{wrapfigure}{L}{0.6\textwidth}
	\begin{minipage}{0.6\textwidth}
 \begin{figure}[H]
     \centering
     \includegraphics[width=0.98\textwidth]{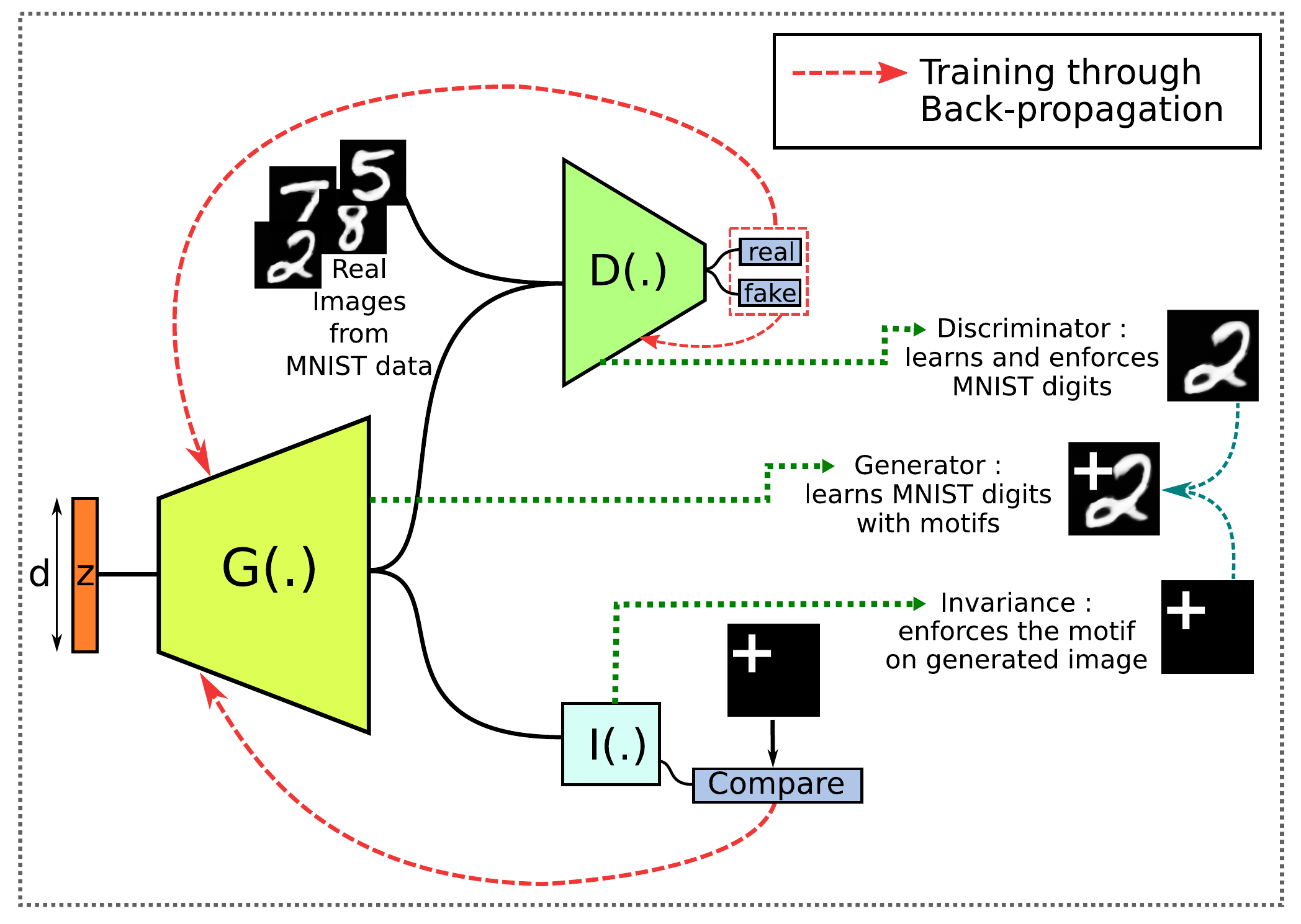}
     \caption{\sl Our proposed InvNet framework introduces novel Invariance checker ($I(.)$) along with traditional generator ($G$) and discriminator ($D$). While the discriminator learns the implicit features of the image through simultaneous training of both $D$ and $G$, our invariance strictly enforces the predefined structure on the generator ($G$) through minimizing invariance loss by updating generator parameters.}
     \label{fig:intro_dia}
 \end{figure}
 \end{minipage}
 \end{wrapfigure}

\textbf{2. \emph{Simulating solutions to PDEs.}} As our primary application, we design InvNet models to generate solutions to partial differential equations, given their coefficients and boundary conditions. For physical systems that are completely described by their governing PDEs, InvNet is entirely \emph{data-free}, i.e., trained solely by enforcing the PDE as a structural invariance. We show that InvNet is also capable of \emph{conditional} generation of solutions, providing flexible user control over both boundary conditions and PDE coefficients. As an example, we solve the classical non-linear time varying Burgers' equation \cite{burgers} in both viscid and inviscid cases, and demonstrate that InvNet provides very competitive results. This marks a significant improvement over recently proposed unsupervised learning methods for solving PDEs, such as~\cite{raissi2017physics1, raissi2017physics2, zhu2019physics}. 

\textbf{3. \emph{Microstructure generation.}} Finally, we also consider a more challenging application in computational material science, which is the problem of simulating (binary) microstructures obeying desired properties \cite{Wodo2012, torquato2013random}. Here, the equations governing the formation of microstructures are typically very complex, but we require that the target microstructures satisfy certain \emph{statistical} invariances. We encode these invariances in the form of moment matching constraints, and train an InvNet model to generate synthetic microstructures adhering to target statistics. We show that InvNet significantly outperforms standard numerical microstructure methods in terms of computational costs. 

\emph{\textbf{Theory}.} We supplement our experimental results with {theoretical analysis}. In particular, we study the InvNet training problem in simple but illuminating special cases. We first perform a (pure) Nash equilibrium analysis for the minimax game in InvNet training, and show that the choice of discriminator architecture plays an influential role in encoding target invariances. We also analyse InvNet training dynamics, and discover that traditional alternating gradient descent updates diverge. To resolve this, we propose training via \emph{extra gradient-descent} and prove its convergence.



\subsection{Related Work}
\label{sec:priorwork}
\textbf{Generative Adversarial Networks.} GANs~\cite{goodfellow2014generative} are a popular approach for modelling real world data distributions. The standard adversarial training approach involves optimizing a mini-max game between a generator and a discriminator defined by an approximate Jensen-Shannon divergence. This necessarily leads to unstable training and requires careful tuning of relevant hyperparameters. Mao \etal  solve the problem of vanishing gradients by optimizing the discriminator with least squares instead of cross-entropy. Arjovsky \etal~\cite{arjovsky2017wasserstein} and Gulrajani \etal~\cite{gulrajani2017improved} propose a more stable version of GAN by modifying the discriminator loss to estimate the Wasserstein-1 distance between the data distribution and the generator output. A major disadvantage for such models is that the inability to control the generator output in any way. Zhao \etal~\cite{Zhao2018BiasAG} study the generalization in GANs.

Conditional GANs~\cite{Mirza2014ConditionalGA} provide a solution by conditioning the generator on categorical labels so as to control the class of outputs that are generated. Chen \etal~\cite{Chen2016InfoGANIR} extend this to allow control over specific semantic parameters such as stroke width in the case of handwritten digits. 

Such generative models therefore offer a mechanism to accurately represent complex data spaces without knowing the entire topology. Consequently, GANs are often used as representations of solution spaces to complex physical and dynamical systems. 

\textbf{Neural networks in physics.} Deep neural networks show the ability to approximate an extremely large set of functions. This property has motivated a significant body of work about using neural networks as functional surrogates in physical systems. This generally involves simulating a specific PDE to create a dataset and further training a neural network to fit the given data. 

Raissi~\etal~\cite{raissi2017physics1, raissi2017physics2} demonstrate the use of a fully connected neural network to generate pointwise solutions to a non-linear PDE using a data sampled from the solution. Pang~\etal~\cite{pang2018fpinns} extend this to fractional PDEs  whle providing theoretical guarantees for convergence. Yang \etal~\cite{yang2018physics} present a similar approach to solve stochastic PDEs using GANs. We note that each of these methods involve using data from either simulated or sampled solutions.

Farimani~\etal~\cite{Farimani2017DeepLT} propose a standard conditional GAN to generate solutions to the standard transport equation where the input is an image representing the initial conditions. Pun~\etal~\cite{pun2018physically} combine a physics based model with a neural network to construct large scale atomistic simulations for material discovery. {De Oliviera} \etal~\cite{de2017learning} propose a DCGAN that additionally learns location based features by using locally connected layers to generate high energy particle simulations. Zhu~\etal~\cite{zhu2019physics, zhu2018bayesian} employ the use a convolutional encoder-decoder architecture along with a conditional FLOW model to surrogate a PDE with stochastic coefficients. Similar to our approach, they rely on the use a physics informed loss to train their model. However, our approach uses adversarial generative models instead of normalizing flows and additionally is flexible enough to allow for variability in both initial conditions as well as the coefficients.

We note that all the above listed approaches rely only on the use of data to enforce the given constraints. Conversely, our architecture builds on known knowledge in the form of invariances, and therefore does not need modified or collected data. We show  this in \Secref{sec:applications} that InvNet learns to solve a PDE for a large variety of initial conditions even when trained for a subset of the same. 

For quantitative comparison, we consider the deterministic surrogate introduced in \cite{zhu2019physics} for solving Burgers' Equation, where input to the framework are samples of input fields that obey certain initial and boundary conditions.

\textbf{Invariance in generative models.} The problem of training a generative model to generate samples from a specific distribution is often solved through the use of data. This approach is extremely useful when the data can not be modeled mathematically. However, for many applications, there exist at least partial mathematical definitions for training data. These mathematical definitions act in place of data, acting as constraints to define the support of the generator distribution.

Stinis~\etal~\cite{Stinis2018EnforcingCF} employ a noisy data training approach with mathematical constraints in order to interpolate and extrapolate on the generator distribution. Contrary to their approach of weakening the discriminator training with noisy inputs, we use an alternating minimization scheme to force the discriminator to respect the invariances.     Jiang \etal \cite{jiang2019spatially} use segmentation masks as constraints to enforce structural conditions to generate face images. Svyatoslav~\cite{svyatoslav18constrainedbNN}, on the other hand, enforce a PDE as a constraint by using a binary neural network as a PDE solver using decision processes. Their approach is restricted to the special case of generating binary images, whereas our algorithm is more general and can enforce any continuous and differentiable invariance.  


\textbf{Dynamics of training GANs.} GANs are known to be notoriously difficult to train, the mini-max optimization problem requiring careful tuning of hyperparameters. As such, there has a been a wealth of recent literature discussing various equilibrium point optimization algorithms~\cite{Heusel2017GANsTB, Qin2018DoGL, Yazici2019TheUE}.  
Li \etal~\cite{Li2018OnTL} demonstrate that first order approximations of discriminator dynamics lead to unstable training and mode collapse.
Mescheder~\etal~\cite{mescheder2017numerics} consider the gradient vector field for the GAN two player game, improving the training algorithm with a regularization. Nagarajan \& Kolter~\cite{nagarajankolterGAN} analyse the standard gradient descent (GD) algorithm as a linear dynamical system and show that under certain assumptions, GD converges to local Nash equilibrium. An interesting approach by Daskalakis \etal~\cite{daskalakis2017training} involves using optimistic mirror decent (OMD), a variant of GD used in optimizing two player games, for linear convergence of GANs to the equilibrium. Liang \& Stokes~\cite{liang2018interaction} additionally study other variants of higher order GD methods and the corresponding interaction between the discriminator and generator dynamics. Mescheder~\etal~\cite{mescheder2018training} present rigourous arguments and experiments studying the convergence of various GAN architectures~\cite{Snderby2017AmortisedMI, Kodali2018OnCA, Roth2017StabilizingTO}. They conclude instance noise and gradient penalty-based training approaches converge locally. Mokhtari~\etal~\cite{mokhtari2019unified} show strong convergence rate results for saddle point problems similar to GANs for extra gradient descent(EGD) and OMD. 

 We build upon Daskalakis~\etal~\cite{daskalakis2017training} and Liang \& Stokes~\cite{liang2018interaction} in \Secref{sec:theory} to rigorously analyse the equilibrium point of our model and further study the convergence of GD and variants. 

\textbf{Microstructure generation.} An entire sub-field in computational material science is devoted to the development of methods for the \emph{simulation} of microstructures~\cite{Ganapathysubramanian2008, Ganapathysubramanian2007, roberts1997statistical} and subsequent quantification~\cite{Ganapathysubramanian2008, Ganapathysubramanian2007}. Here, microstructure realizations are synthesized that satisfy certain target statistical properties of the material distribution. Several strategies were developed for microstructure generation using both analytical approaches and optimization approaches. Examples of such methods include Gaussian random fields~\cite{roberts1997statistical}, optimization-based methods~\cite{torquato1998reconstructing}, multi-point statistics~\cite{Feng2018}, and layer-by-layer reconstruction~\cite{tahmasebi2012multiple}.These statistical properties could be scalars (such like total volume fraction of a material) or more complex functions (like 2-point correlations and other material statistics)~\cite{torquato2013random}. Recent advances also involve the generative modeling techniques~\cite{sanchez2018inverse}, however, those largely rely on the availability of training datasets.

For our experiments in generating microstructures, we use the Binary 2D microstructures dataset~\cite{pokuri_balaji_sesha_sarath_2019_2580293} based on Cahn-Hilliard equation~\cite{cahn1958free} for training and testing.

\begin{table}[!t]
	\caption{\sl Qualitative comparison of InvNet with existing approaches across different applications}
	\label{tab:compint}
	\small{
		\centering
		\newcommand{\tabwid}{1.10cm}
			\renewcommand{\arraystretch}{1}
			\begin{threeparttable}
\begin{tabular}{p{2.5cm}>{\centering\arraybackslash}p{1.9cm}>{\centering\arraybackslash}p{1.15cm}>{\centering\arraybackslash}p{1.0cm}>{\centering\arraybackslash}p{0.9cm}>{\centering\arraybackslash}p{1.0cm}>{\centering\arraybackslash}p{\tabwid}>{\centering\arraybackslash}p{1.9cm}>{\centering\arraybackslash}p{1.00cm}}
	\toprule
	Application$\rightarrow$ & \multicolumn{2}{c}{\small{Encoding motifs}}  & \multicolumn{3}{c}{\small{Solving PDE}} &  \multicolumn{3}{c}{\small{Microstructure generation}} \\ 
	\cmidrule{1-1} \cmidrule(lr){2-3} \cmidrule(lr){7-9} \cmidrule(lr){4-6}
	Model $\rightarrow$ & WGAN-GP\cite{gulrajani2017improved} & \textbf{InvNet}  & PINN~\cite{raissi2017physics1} & PCS~\cite{zhu2019physics} & \textbf{InvNet}  & Numerical method~\cite{Wodo2012} & WGAN-GP\cite{gulrajani2017improved} & \textbf{InvNet} \\ 
	\cmidrule{1-1} \cmidrule(lr){2-3} \cmidrule(lr){7-9} \cmidrule(lr){4-6}
Requires training & Yes  & Yes   & Yes & Yes  & Yes & No  & Yes  & Yes\\ 
\cmidrule{1-1}
Requires data \emph{to learn invariances} & Yes & \textbf{No}   & Yes & No & \textbf{No} & No  & Yes   & \textbf{No}\\ 
\cmidrule{1-1}
Can be trained without discriminator & Yes & Yes & No & No$^+$ & Yes & - & Yes & No \\ 
\cmidrule{1-1}
Supports conditional model for invariances & No  & \textbf{Yes}  & No  & No  & \textbf{Yes}& - & No  & \textbf{Yes} \\ 
\cmidrule{1-1}
Generation rate (samples/unit time) & High & High    & High & High & High  & Low  &  High   & High \\ 
\bottomrule
\end{tabular}
\begin{tablenotes}\scriptsize
	\item[\scriptsize{+}] decoder in this case
\end{tablenotes}
\end{threeparttable}}
\end{table}
\section{Proposed Model: InvNet}
 \label{sec:methods}

Consider a data distribution $\sP_{\text{data}}$ defined over set $\gD \subseteq \R^d$, and a list of differentiable \emph{invariance} functions $\R^{d} \rightarrow \R: I_i(\cdot), i = 1,2,..,r$. The aim of InvNet is to generate new samples $\rvx$ from $\gD$ that satisfy the expression $I_i(\rvx) = 0,~\forall i = 1,2,..,r$. We define our generator to be a function $G_{\theta}: \R^{k} \to \R^{d}$ parameterized by $\theta$. Let $\rvz$ represent a $k$-dimensional latent input vector. 
In the standard GAN setup~\cite{goodfellow2014generative}, the generator is trained by posing a two-player game between a generator ($G$) and a discriminator ($D$), where the discriminator is a function $D_\psi:\sR^d \to \sR$ parameterized  by $\psi$. 
Using the notation of \cite{mescheder2018training}, the training objective for GAN is described by:
\begin{equation}
    L(\theta, \psi) = \E_{\rvx \sim \sP_{\text{data}}}\left[f(D_\psi(\rvx))\right] + \E_{\rvz \sim \sP_z}\left[f(-D_\psi(G_\theta(\rvz)))\right],
    \label{eq:ganloss}
\end{equation}
for some monotonic function $f:\sR\to\sR$ and $\sP_z$ being a known distribution. We focus on Wasserstein GAN~\cite{arjovsky2017wasserstein, gulrajani2017improved}, where $f(t)=t$, while noting that our approach below extends to general (differentiable) $f$ functions \emph{mutatis mutandis}. 
In order to encode the invariances, we propose the following minimax game to train our InvNet model: 
\begin{align}
     \label{eq:inv_net}
     \min_{\theta} \max_{\psi} \bar{L}(\theta, \psi),~\text{where}~\bar{L}(\theta, \psi)&=
      L(\theta, \psi) + \mu \sum_{i=1}^n \E_{z} \left[I_i(G_\theta(\rvz))\right] := L(\theta,\psi) + \mu L_I(\theta). 
\end{align}

We solve the minimax game in a fashion similar to GAN training; we alternately adjust the generator parameters $\theta$ and the discriminator parameters $\psi$ via gradient updates. However, due to the presence of the additional invariance term in $\bar{L}(\theta,\psi)$, we find in practice that a three-way update rule works well: a GAN-like update of $\theta$ via gradient steps of $L(\theta,\psi)$ keeping $\psi$ fixed; a GAN-like update of $\psi$ via gradient steps of $L(\theta,\psi)$ keeping $\theta$ fixed; and an update of $\theta$ via gradient steps of $L_I$. See Alg~\ref{alg:aio}.
	\begin{algorithm}[H]
		\begin{small}
			\caption{Alternating Optimization for InvNet}
			\label{alg:aio}
			\begin{algorithmic}[1]
				\Require Set learning rates, termination conditions. 
				\While {$L_I$ large and $\theta$ has not converged}
				\For {$l \gets 1\ \text{to}\ N_G$}
				\State  $\theta \gets \theta - \eta_G\nabla_\theta \bar{L}$ \Comment{Generator update}
				\EndFor
				\For {$m \gets 1\ \text{to}\ N_D$}
				\State $\psi \gets \psi + \eta_D \nabla_\psi \bar{L}$ \Comment{Discriminator update}
				\EndFor
				\For {$n \gets 1\ \text{to}\ N_I$}
				\State $\theta \gets \theta - \eta_D \nabla_\theta L_I$ \Comment{Generator update}
				\EndFor
				\EndWhile
			\end{algorithmic}
		\end{small}
	\end{algorithm}


{\textbf{Role of the input vector $\rvz$}.} Unlike the standard GAN setting, we define $\rvz$ as the concatenated vector $[\tilde{\rvz}, \rvc]^T$ with $\tilde{\rvz}$ referring to a random vector sampled from a known distribution while $\rvc$ is a deterministic vector that parameterizes the invariances. This way, $\rvz$ can be used to model both stochastic and deterministic components of the data. Moreover, $\rvc$ gives us additional tuning knobs, allowing a broader range of applications for the InvNet model; we elaborate below.

{\textbf{Variations of alternating optimization}.} As mentioned above, we optimize our multi-objective formulation by alternately optimizing over the three sub-components of $\bar{L}$, as presented in Alg.~\ref{alg:aio}. Our choice of using alternating optimization is informed by insights in recent work~\cite{daskalakis2017training, mokhtari2019unified, liang2018interaction} that show that regular gradient descent for minimax games diverges, while methods that take intermediate gradient steps, such as \emph{extra-gradient descent}, converge to stable Nash equilibria. 
\Secref{sec:theory} provides a rigorous analysis of extra-gradient descent for training InvNets in special cases.


{\textbf{When the invariances perfectly define the target distribution}.} In cases where the set of invariances, $I_i$, are necessary \emph{and} sufficient conditions for specifying $\gD$, 
the invariances themselves can be interpreted as serving the role of the \emph{optimal} discriminator. This has two major implications: (i) $D_\psi$ is constant, Lines 5-7 of the algorithm are no longer needed, and \Eqref{eq:inv_net} is a regular minimization problem that can be stably solved; (ii) more importantly, since the gradients of $\bar{L}$ no longer depends on the data distribution $\mathbb{P}_{data}$, InvNet training can be performed in a {\it data-free manner}. This property is crucial in modeling complex physical systems that are fully described through their governing PDEs. We present one such application in \Secref{sec:applications}.

\section{Applications of InvNet}
\label{sec:applications}


\subsection{Planting Motifs in Images}
\label{subsec:motif}
We start with the stylized problem of planting a predefined 
pattern, or motif, at a fixed location in synthetically generated image data. 
For example, such a scenario might arise if a logo (or signature, or watermark) is required to be present in all synthetic images. 

There are two obvious approaches to solve this problem: simply blend/paste the motif onto any output image synthesized by a regular generative model; or use a standard GAN model with a new modified training dataset, where each training image has the motif blended in. However, modifying output images as well as input training data could be non-trivial. 

We propose a novel approach via InvNet. Consider a problem of synthesizing images containing a `plus' symbol ($+$) as a motif centered at a fixed location. The motif can be encoded via a image-sized binary mask $\mM_{+}$ with a black background and a white `plus' sign. We define an invariance-loss based on the above structural description as follows:
\begin{align}
L_I(G_\theta(z)) = \| \gP_{\Omega}(G_\theta(z)) - \mM_{+} \|^2_F \, ,
\label{eq:motifinvar}
\end{align}
where $\gP_{\Omega}$ is the projection (indicator) function for the motif. 

  \newcommand{\wc}{0.09}
 \begin{figure}[t]
    	\renewcommand{\arraystretch}{1.2}
    \begin{tabular}{c c c}
    \raisebox{0.2\height}{\includegraphics[width=0.24\linewidth,cframe=ube 1pt]{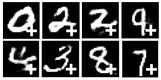}} & \raisebox{0.2\height}{\includegraphics[width=0.24\linewidth,cframe=amber 1pt]{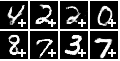}} &

    \multirow{1}{*}[6em]{\resizebox{0.40\linewidth}{!}{
    \begin{tikzpicture}
          \begin{axis}[axis background/.style={fill=seabornback, fill opacity=1},
        ylabel={Discriminator loss},
        grid style={line width=.1pt, draw=white},
        major grid style={line width=.1pt,draw=white},
        minor tick num=1,
        xtick = {0,50,100,150,199},
     xticklabels  = {0,7500,15000,22500,30000},
        grid=both,
        xlabel= {Training iterations},
        legend cell align=left,
        legend style={at={(1,0.25)}}]
\addplot[ultra thick,color=ube] table[x=index, y=standard_D, col sep=comma]{data/motif_disc_loss_compare_resampled.csv};
\addplot[ultra thick,color=amber] table[x=index, y=special_D, col sep=comma]{data/motif_disc_loss_compare_resampled.csv};
  \legend{{Standard discriminator},{Modified discriminator}}
 \end{axis}
 \end{tikzpicture}}} \\
 \raisebox{0.2\height}{\includegraphics[width=0.24\linewidth,cframe=ube 1pt]{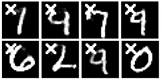}} & \raisebox{0.2\height}{\includegraphics[width=0.24\linewidth,cframe=amber 1pt]{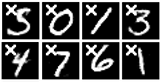}} & \\
    \small{(a)} & \small{(b)} & \small{(c)} \\
    \end{tabular}
    \setlength{\tabcolsep}{1pt}
	\renewcommand{\arraystretch}{0.5}
	\begin{tabular}{ccccccccccc}
			\raisebox{0.8em}{\small{(d)}} &
			\includegraphics[width=\wc\linewidth]{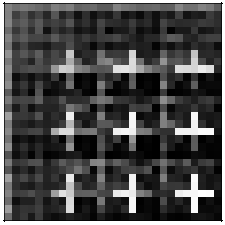} &
			\includegraphics[width=\wc\linewidth]{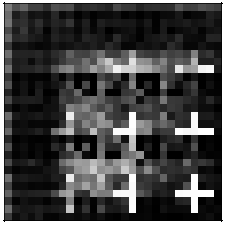} &
			\includegraphics[width=\wc\linewidth ]{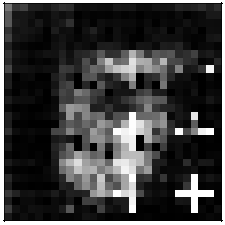} &
			\includegraphics[width=\wc\linewidth ]{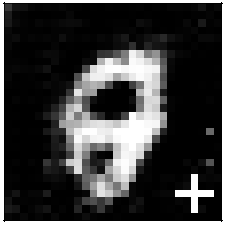} &
			\includegraphics[width=\wc\linewidth ]{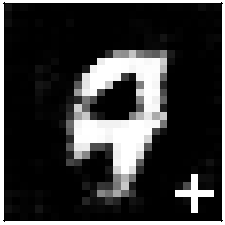} &
			\includegraphics[width=\wc\linewidth ]{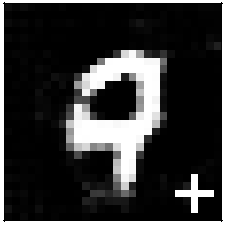} &
			\includegraphics[width=\wc\linewidth]{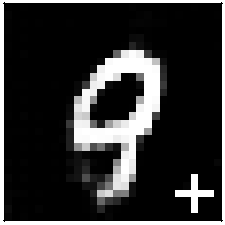} &
			\includegraphics[width=\wc\linewidth]{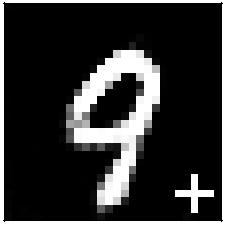} &
			\includegraphics[width=\wc\linewidth]{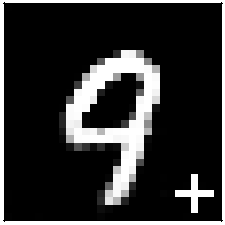} &
			\includegraphics[width=\wc\linewidth]{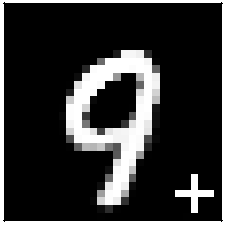} 
    \end{tabular}
    \caption{\sl Images generated by the InvNet for planting a motif on MNIST images with a `plus' ($+$) and `cross'$(\times)$ as the motif with (a) standard WGAN-GP discriminator; (b) modified discriminator with complimentary projection. (c) Loss curve shows the variation of discriminator loss for corresponding cases. (d) Evolution of generated images during training across 25 epochs (25000 iterations, batch size 50). The invariant motif is learnt quickly due to alternating optimization.}
    \label{fig:motif}
\end{figure}
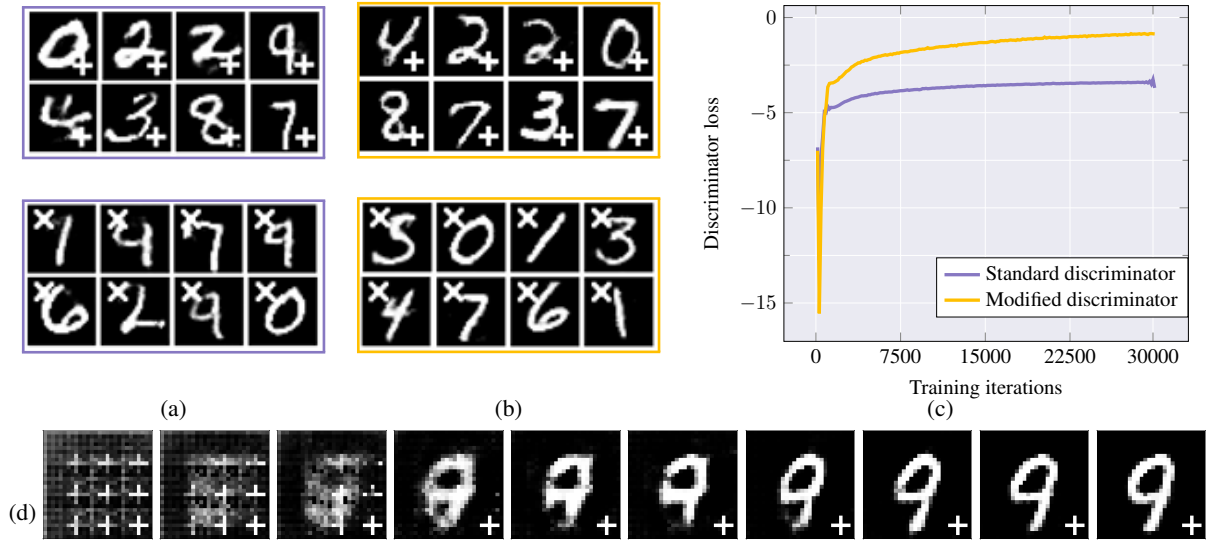
We train an InvNet model using the MNIST dataset, coupled with the motif invariance as defined above. Contrary to typical GAN models, InvNet learns to model both the underlying data, i.e., the digits while additionally learning to generate the required motifs. An analogous experiment can be conducted for a different motif (say a $\times$ symbol) with similar high-quality results; see Fig.~\ref{fig:motif}(a,b). In Fig.~\ref{fig:motif}(d), the evolution of generated images during the training process is visualized; the motif is quickly encoded while learning the implicit features of the data takes several epochs.

We observe an interesting phenomenon: when the generator starts to learn to encode the motif invariance, the discriminator quickly identifies fake images by detecting presence of the motif (since the original MNIST images did not contain the motif at all). This results in poorer quality synthetic samples. We alleviate the problem by instructing the discriminator to \emph{ignore} the pixels corresponding to the support of motif; mathematically,  we apply an orthogonal projection operator $\gP_{\Omega^c}$ to each generated image that is fed as input of discriminator. 
Thus, the discriminator $D_\psi(.)$ learns a distribution over $\gP_{\Omega^c}(\rvx),$, i.e., the space of pixel locations not in the motif. 

Fig.~\ref{fig:motif}(b) shows generated images for the modified discriminator case. Note the improved discriminator loss (orange curve) in this case. We study this phenomenon in greater detail in our theoretical analysis in Section~\ref{sec:theory} and rigorously prove that in special cases, modifying the discriminator is necessary for convergence to a (pure) Nash equilibrium. 


\subsection{Generating Solutions for PDEs}
\label{subsec:burgers}

\newcommand{\wb}{0.16\textwidth}
\newcommand{\wbs}{0.16\textwidth}
\begin{figure}[t!]
	\centering
	\setlength{\tabcolsep}{1pt}
	\renewcommand{\arraystretch}{1.2}
	\begin{tabular}{>{\centering\arraybackslash}p{1.3cm} c c c c c c} 
	    {} & \small{\(\nu=0.0\)} & \small{\(\nu=0.0\)} & \small{\(\nu=0.001\)} & \small{\(\nu=0.002\)} & \small{\(\nu=0.05\)} & \\
	     {} & \small{\(c=2.0\)} & \small{\(c=4.0\)} & \small{\(c=4.5\)} & \small{\(c=4.5\)} & \small{\(c=4.0\)} & \\
		\raisebox{0em}{\begin{sideways}\small{~~~~~~InvNet}\end{sideways}} & \begin{sideways}~~~~~~~$x \rightarrow$\end{sideways} 
		\includegraphics[trim=4pt 4pt 50pt 4pt,clip,width=\wbs]{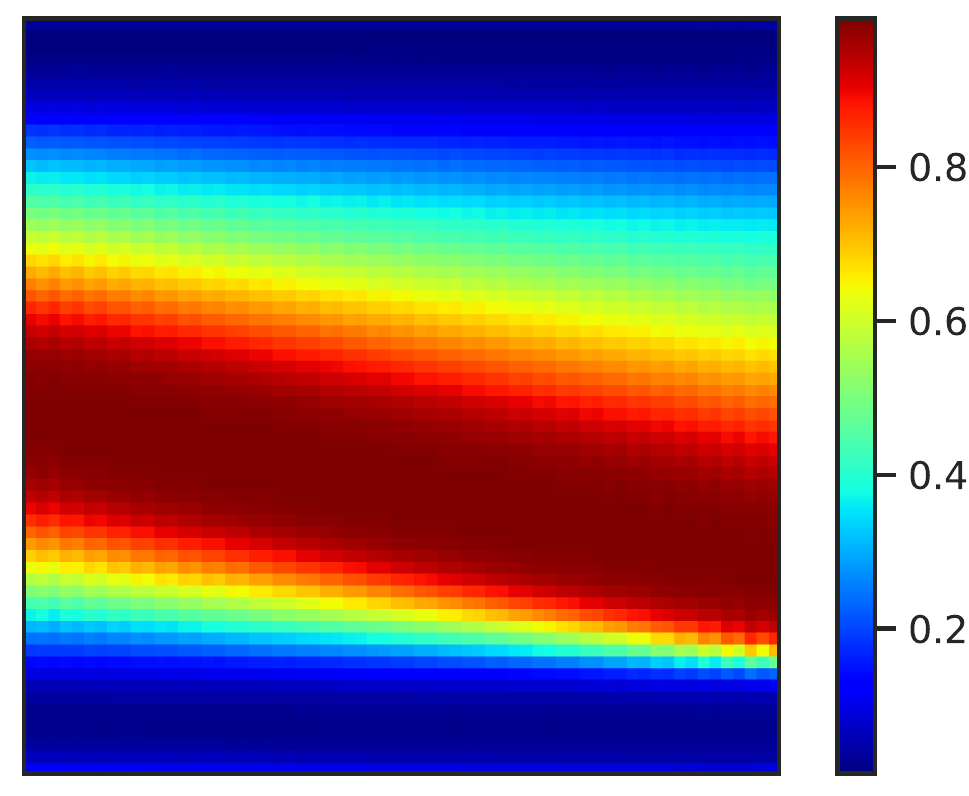} &
		\includegraphics[trim=0 0 50pt 0,clip,width=\wbs]{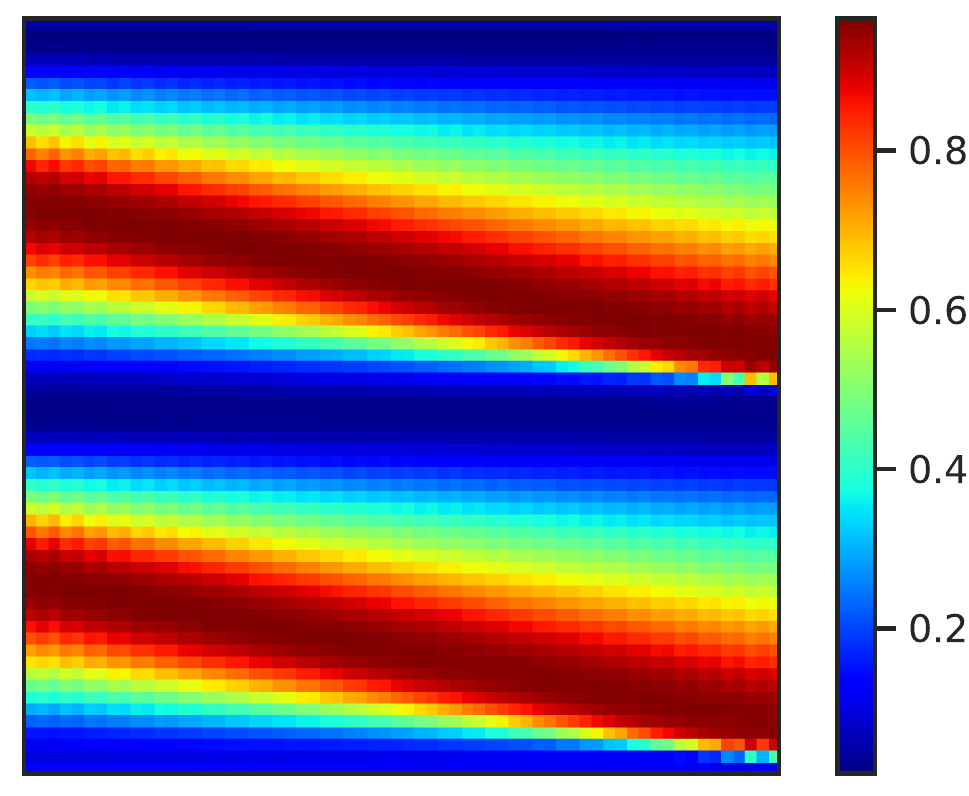} & 
		\includegraphics[trim=0 0 50pt 0,clip,width=\wbs]{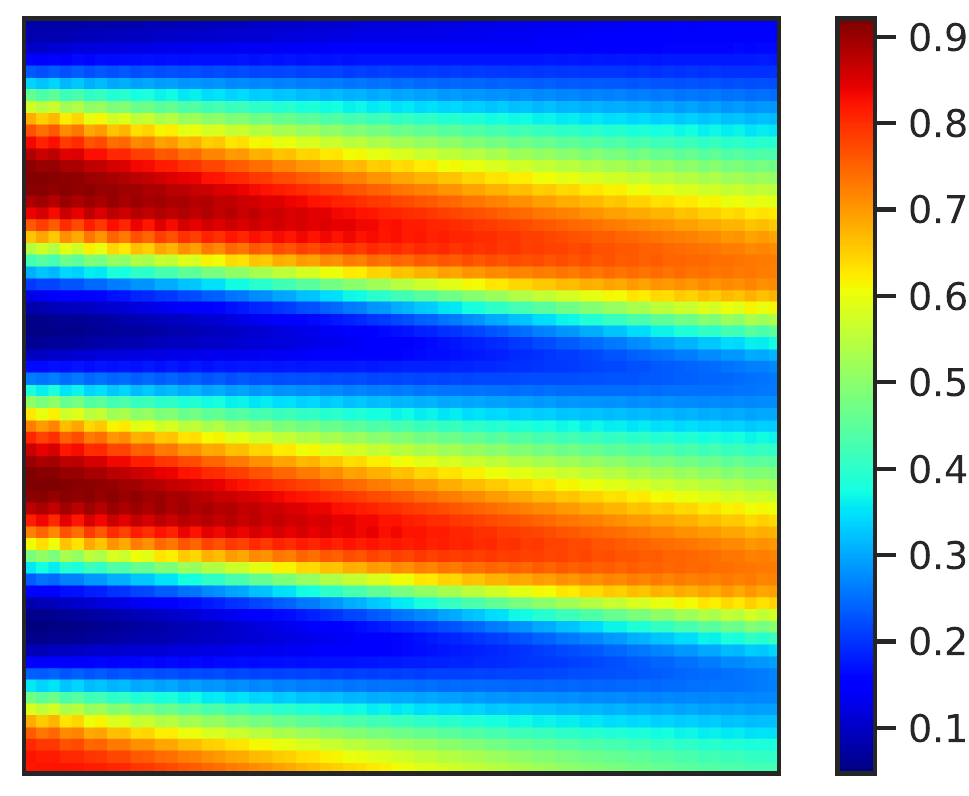} & 
		\includegraphics[trim=0 0 50pt 0,clip,width=\wbs]{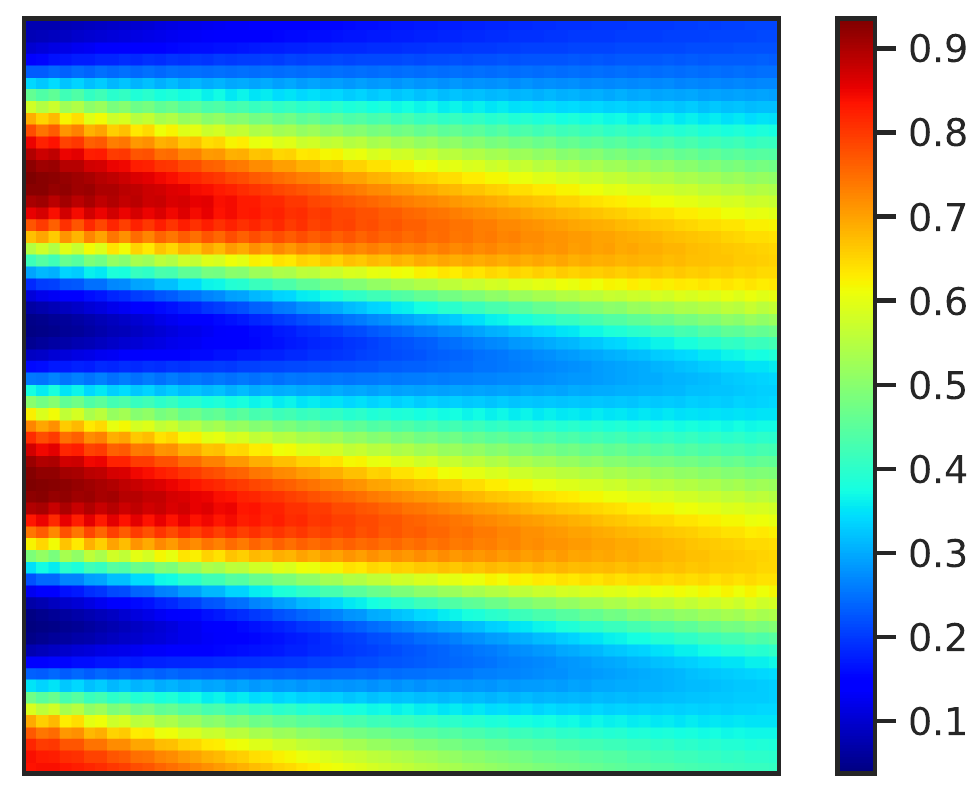} &
		\includegraphics[trim=0 0 0 0,clip,width=0.156\textwidth]{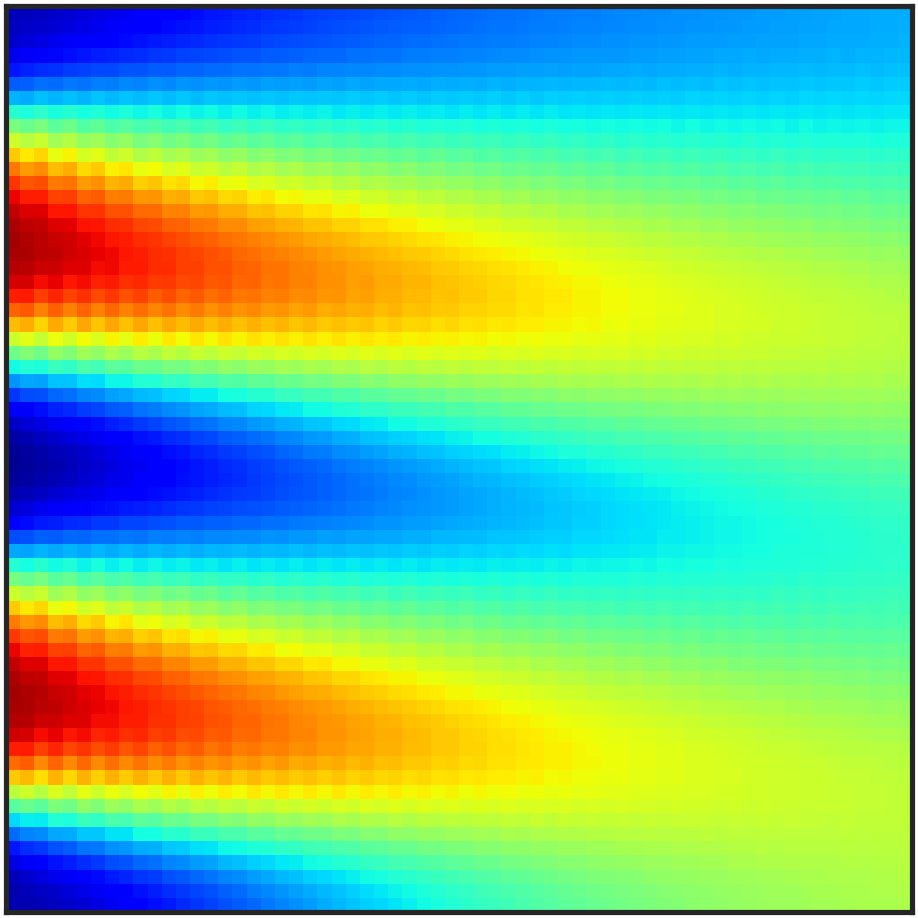} &
		\multirow{4}{*}[1em]{\includegraphics[trim=23em 0 0 0, clip,width=0.06\textwidth]{fig/burgers/new/viscid_35_46-eps-converted-to.pdf}} \\
	    \raisebox{0em}{\begin{sideways}\scriptsize{~~Numerical method}\end{sideways}} & \begin{sideways}~~~~~~~$x \rightarrow$\end{sideways} 
		\includegraphics[trim=0 0 0 0,clip,width=\wbs]{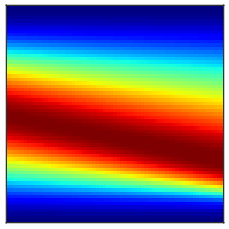} & 
		\includegraphics[trim=0 0 0 0,clip,width=\wbs]{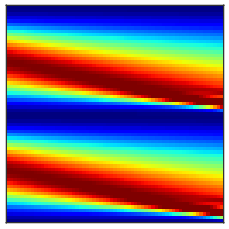} & 
		\includegraphics[trim=0 0 0 0,clip,width=0.155\textwidth]{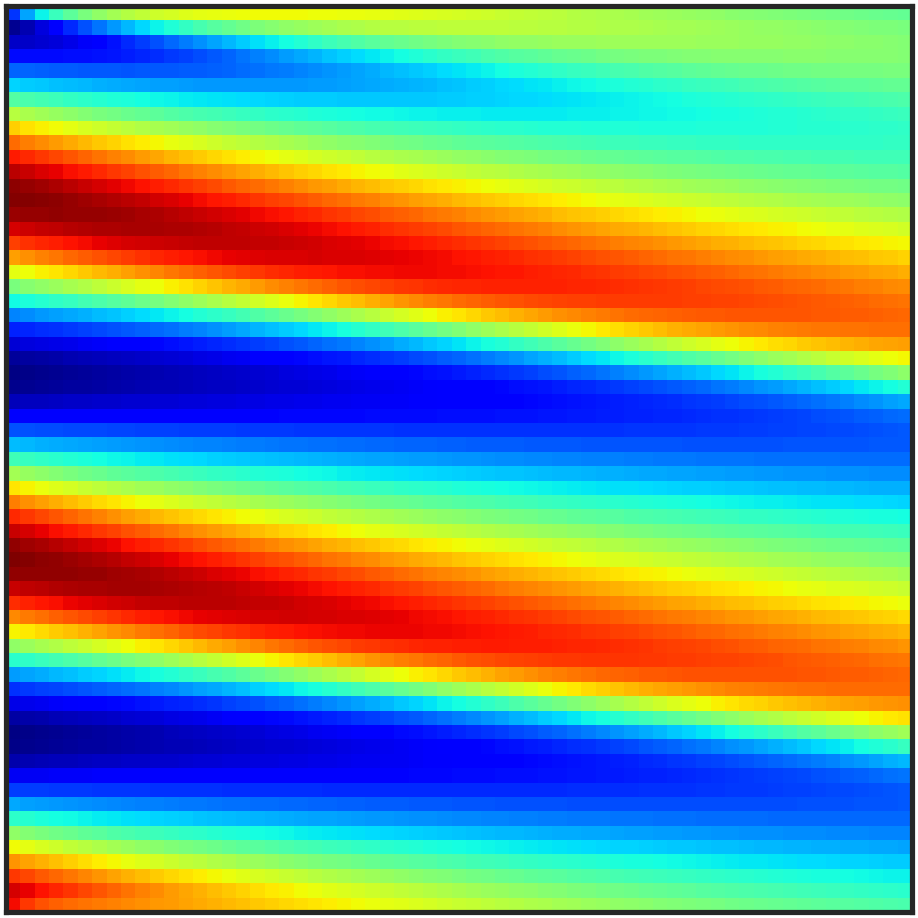} & 
		\includegraphics[trim=0 0 0 0,clip,width=0.155\textwidth]{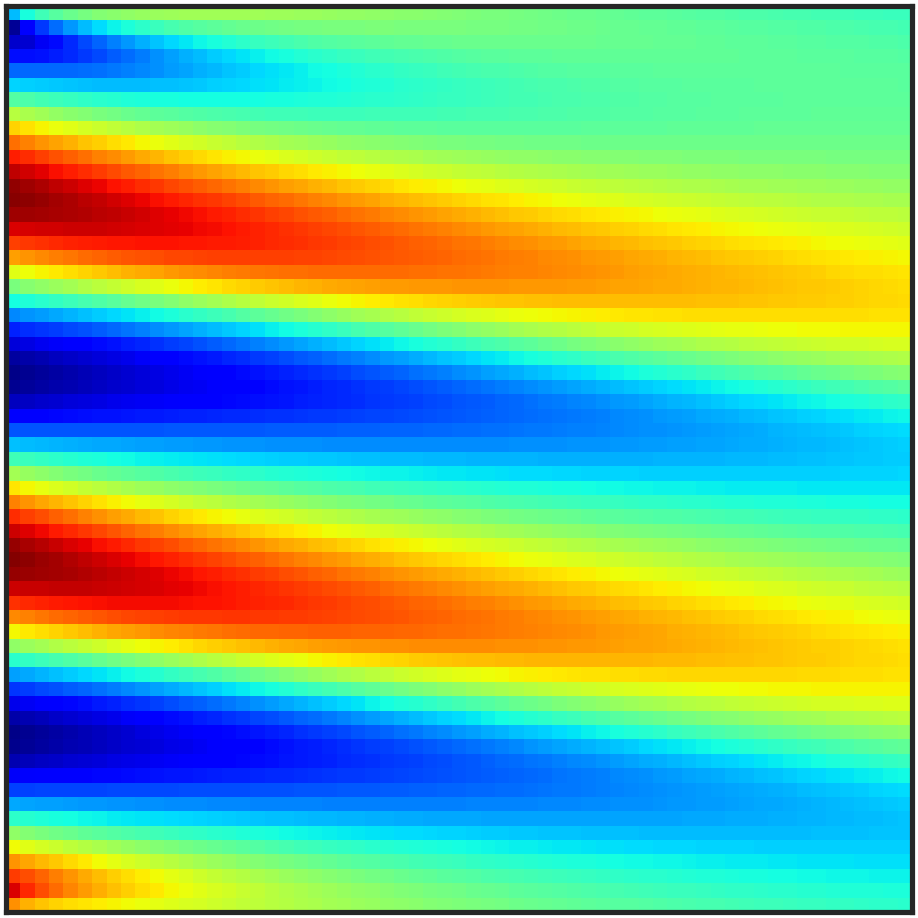} &
		\includegraphics[trim=0 0 0 0,clip,width=0.156\textwidth]{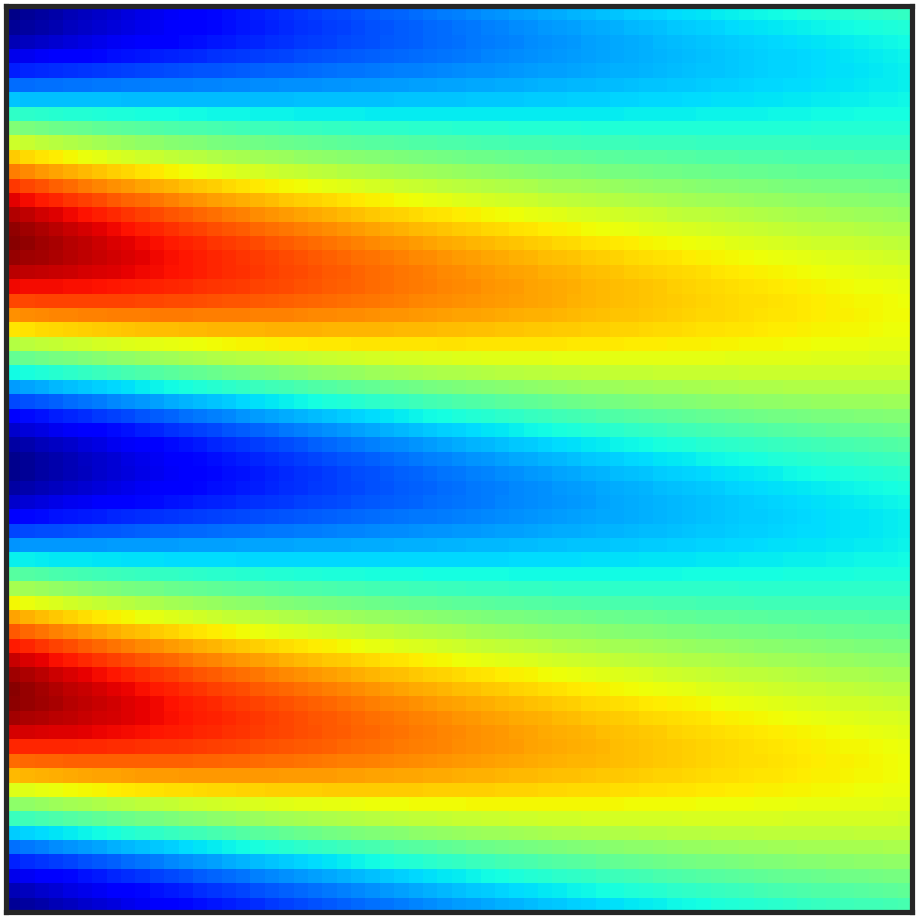} &
		\\
        \raisebox{0em}{\begin{sideways}\small{~~~~~~Residual}\end{sideways}} & \begin{sideways}~~~~~~~$x \rightarrow$\end{sideways}
		\includegraphics[trim=0 0 50pt 0,clip,width=\wbs]{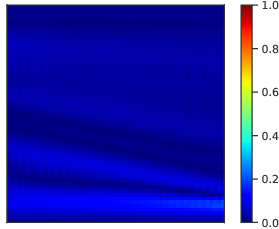} & 
		\includegraphics[trim=0 0 50pt 0,clip,width=\wbs]{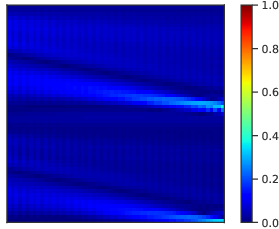} & 
		\includegraphics[trim=0 0 50pt 0,clip,width=0.156\textwidth]{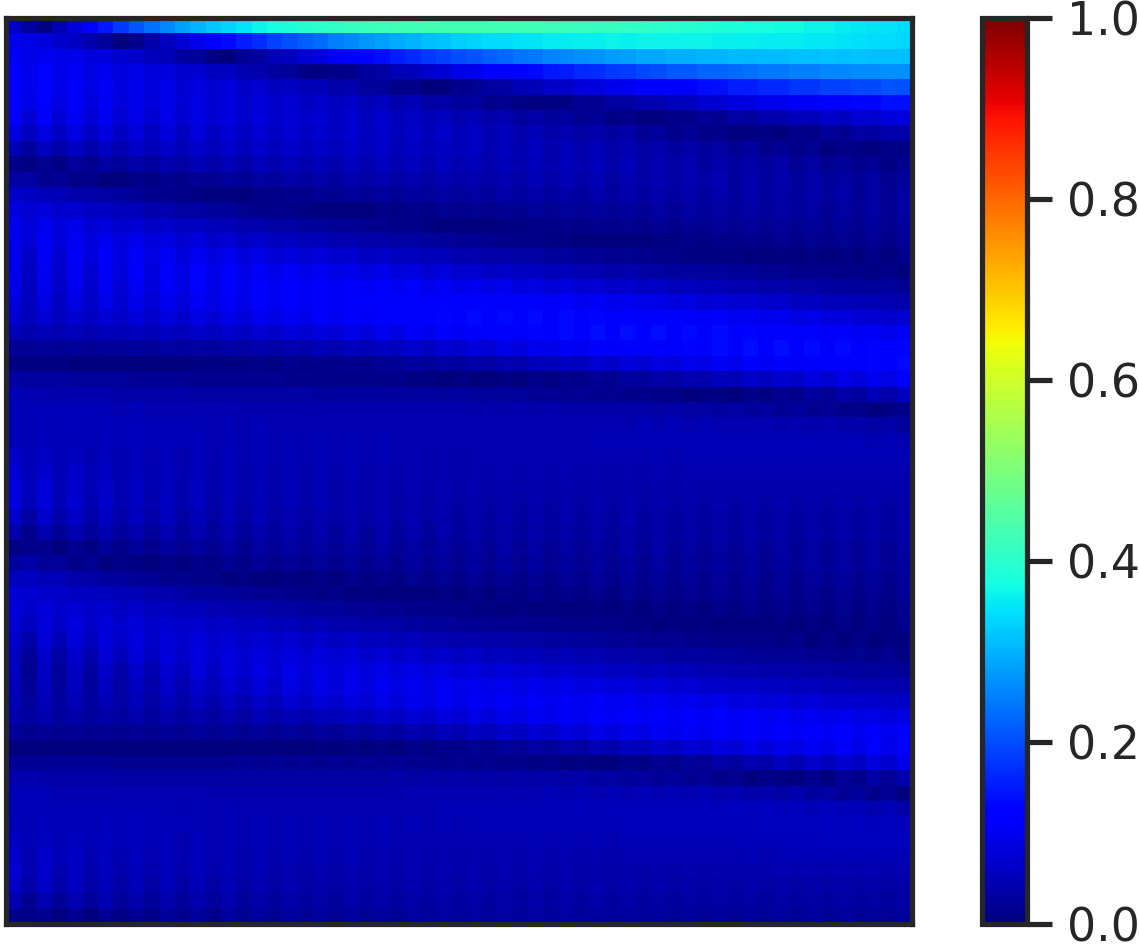} & 
		\includegraphics[trim=0 0 50pt 0,clip,width=0.156\textwidth]{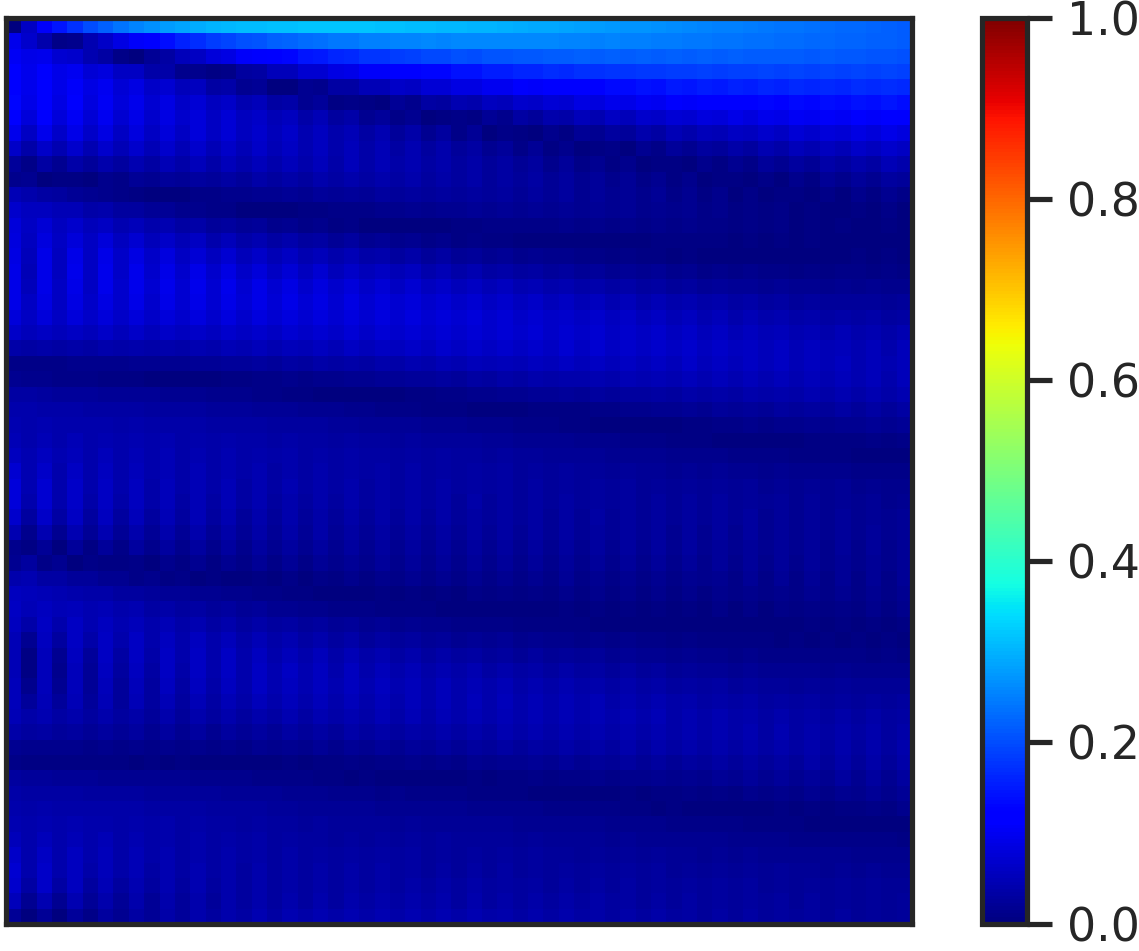} &
		\includegraphics[trim=0 0 50pt 0,clip,width=0.156\textwidth]{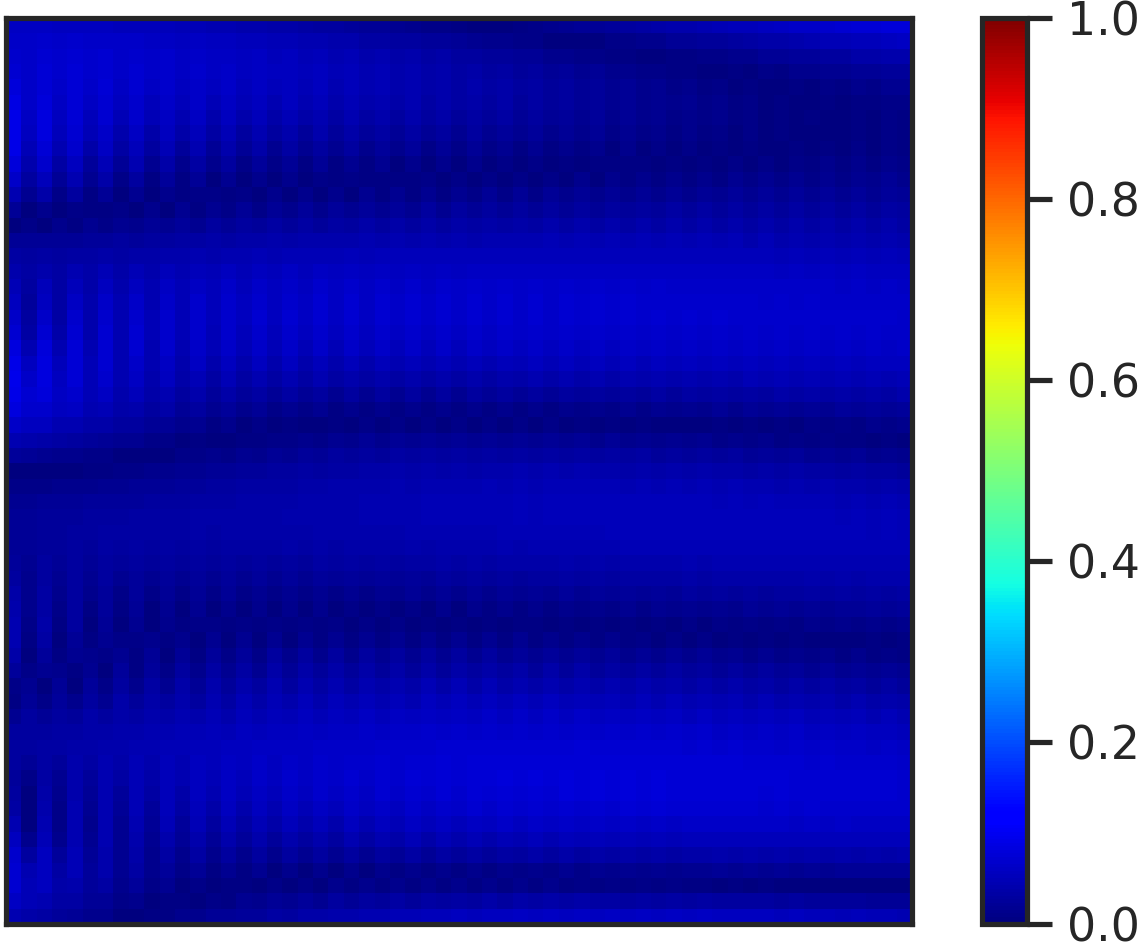}
		\\
			& \raisebox{0.01\textheight}{$t \rightarrow$} &\raisebox{0.01\textheight}{$t \rightarrow$} &\raisebox{0.01\textheight}{$t \rightarrow$} &\raisebox{0.01\textheight}{$t \rightarrow$} & \raisebox{0.01\textheight}{$t \rightarrow$} & \\
	Relative $\ell_2$ error & $0.073$ & $0.095$ & $0.147$ & $0.117$ & $0.071$
	\end{tabular}
	\caption{\sl Example of solutions generated using the InvNet for a family of boundary conditions. Row (a) shows solutions generated by InvNet. Row (b) are numerical solutions calculated by explicit Euler methods. Row (c) represents the residual between the two. Column (1) explicitly shows that our model generalizes even for an unseen initial condition, $c=2.0$.
	 For the inviscid case (Cols. (1) \& (2)), the wave, defined by the boundary condition (1st column of the image) propagates in time without any diffusion. In case of the viscid equation (Cols. 3, 4 \& 5), the wave energy diffuses across the image, as evidenced by the decaying amplitude. The relative error with respect to the numerical solution calculated using explicit Euler methods is comparable to \cite{zhu2019physics}.}
	\label{fig:burgerssoln}
\end{figure}

We demonstrate that the solution set of partial differential equations (PDEs) describing dynamics of physical systems can be accurately recovered using an InvNet model. 
Given the parametric form of a system of PDEs, its solution set is governed by the given physical constants (parameters) together with any boundary conditions. 
Recently, data-driven methods have been proposed to successfully employ deep generative models as fast surrogates to traditional PDE solvers~\cite{zhu2019physics, zhu2018bayesian}. However, InvNet provides an alternate, data-free approach to solving PDEs. The key idea is that since the PDE itself serves as the optimal discriminator, Lines 5-7 of Alg.\ \ref{alg:aio} are redundant and no training data is required.  

We demonstrate this via a simple non-linear PDE called \emph{Burgers' Equation}~\cite{burgers}. We use a conditional input $\rvc$ to InvNet to control the solution set of Burgers'. Therefore, unlike~\cite{zhu2019physics}, a \emph{single} well-trained InvNet can generate solutions corresponding to a variety of boundary conditions as well as varying physical parameters. Before we describe our approach, we provide a quick PDE primer.

\textbf{Inviscid Burgers' equation.} This is a non-linear PDE encountered in fluid mechanics and nonlinear acoustics. The inviscid form of the Burgers' equation, defined in \Eqref{eq:burgers}, assumes a non-diffusive fluid through which a wave with initial state $f_i$ is passed. Let $\rmU(x,t)=[\rvu_0, \rvu_1,\dots,\rvu_n]$ be a particular solution of the Burger's equation. Then,
\begin{eqnarray}
     \label{eq:burgers}
        \rmU_{,t} + \rmU_{,x} = 0; 
    \label{eq:init_cond}
    ~~\rmU(x, 0) = f_i(x) ,
\end{eqnarray} where the partial derivative with respect to $x$ (or $t$) is represented as $\rmU_{,x}$ (or $\rmU_{,t}$).

Suppose we model the field $\rmU(x,t)$ as an image (where rows correspond to space and columns correspond to time; see Fig~\ref{fig:burgerssoln}). We train a variant of InvNet that generates solutions, $G_\theta(\rvz)$ to \eqref{eq:burgers} for a given boundary condition as input. 
The boundary condition is a structural invariance that we enforce by minimizing the $\ell_2$ loss between the boundary of the generated solution $\rmU(x, 0)=\gP_\Omega(G_\theta(\rvz))$ and the given boundary $\rvb$. 
In our experiments, we set $\rvb$ as a (discretized) raised-cosine function parameterized on $c$. We sample $\rvb$ uniformly from set $B$ given as:
\begin{equation}
    B = \{ x | x = \frac{1}{2} \left( 1 - \cos(2\pi \rvx c/d) \right),~~ c \in [3.0, 6.0]\}\}.
    \label{eq:init_param}
\end{equation}

Our InvNet minimizes (a weighted combination of) the following losses: 
\begin{eqnarray}
    \label{eq:burgers_loss_gen}
    L_G = \E_z \left[|| G_\theta(\rvz)_{,t} +  G_\theta(\rvz) \odot G_\theta(\rvz)_{,x} ||_2^2\right],~~L_I = \E_z \left[ || \gP_\Omega(G_\theta(\rvz)) - \rvb ||^2_2\right]
\end{eqnarray}
We calculate the partial derivatives by convolving the generated image with the directional Sobel operators~\cite{sobel19683x3}
, reducing the border effects by replication. 
We show in Fig.~\ref{fig:burgerssoln} that InvNet successfully learns to generate solutions for the family of boundary conditions given by $B$ in~\Eqref{eq:init_param}.

\textbf{Viscid Burgers' equation.} We extend the above algorithm to generate solutions for a \emph{family} of PDEs, with a physical scalar parameter indexing each PDE. We consider the family of viscid Burger's equation represented in \Eqref{eq:viscid} with the viscosity term, $\nu$:
\begin{equation}
\rmU_{,t} + \rmU \odot \rmU_{,x} = \nu \rmU_{,x,x} .
\label{eq:viscid}
\end{equation}
Similar to the inviscid case, we provide both the boundary condition $\rvb$ and the coefficient of viscosity $\nu$ as input to the generator. Our input vector becomes $\rvz = \rvc = [\rvb, \nu]$, where $\rvb$ is sampled uniformly from $B$ (\Eqref{eq:init_param}), and $\nu$ is sampled randomly from the set $\text{Unif}[0.001, 0.05]$. Fig.~\ref{fig:burgerssoln} depicts the solutions corresponding to different combinations of $\rvb$ and $\nu$.


We compare our InvNet-based surrogate solutions to numerical solutions (computed using explicit Euler methods) for both the viscid and inviscid cases in Fig.~\ref{fig:burgerssoln}, and observe that our model is accurate.
Moreover, InvNet shows effective generalization by generating solutions for boundary conditions that were \emph{not} used during training.

\noindent\textbf{Comparison with PDE surrogate Encoder Decoder Networks} 
We compare the performance of InvNet with the convolutional Encoder-Decoder (ED) method proposed in ~\cite{zhu2019physics}. Zhu~\etal~propose an encoder-decoder architecture that does not use labelled data (or explicit PDE solutions) but takes in a field with an initial condition as input and optimizes over a numerical form of the PDE and boundary conditions to generate solutions. Conversely, we use a more general generative architecture that can be extended to solve families of PDEs for a large class of initial conditions. We adapt their approach for solving the viscid Burgers' equation for the same range of coefficients and initial conditions. A point worth noting here is that they optimize over the relaxed Lagrangian form of the objective function. On the contrary, we hold to our alternating optimization setting.

We analyse the solutions generated InvNet and ED networks for a range of initial conditions. We train an InvNet to solve a family of viscid equations with parameters, $\nu \sim \text{Unif}(0.001, 0.05)$. For a fair comparison, we additionally train ED networks for sampled values of \(\nu\) for the same initial conditions of the form in \Eqref{eq:init_param}. Note that a single instantiation of ED networks solves a particular PDE with a particular initial condition.  

Tab.~\ref{t:zhucomp} shows relative $\ell_2$ errors and images generated by both approaches. We show comparable performance with a single trained InvNet when contrasted with solutions from the corresponding ED networks.

\begin{table}[!t]
    \centering
    \caption{Relative $\ell_2$ error of solutions generated by InvNet compared to that of ED networks. The error is calculated with respect to numerical solutions. Note that a single trained InvNet shows comparable performance to a range of trained ED networks.}
    \vspace{2.5mm}
    \begin{tabular}{c c c c}
    \toprule
         Viscosity coefficient $\nu$ & Frequency of boundary conditions ($c$) & InvNet & ED~\cite{zhu2019physics} \\
         \midrule
         \multirow{2}{*}{0.002} & 2.0 & 0.08  &  0.04 \\
                            {}  & 4.0 & 0.13 &  0.08\\
                            {}  & 5.5 & 0.29  & 0.30 \\
                            {}  & 6.0 & 0.20 &  0.15\\
         \midrule[0.5pt]
         \multirow{2}{*}{0.006} & 2.0 & 0.08  & 0.04 \\
                            {}  & 4.0 & 0.10 &  0.14\\
                            {}  & 5.5 & 0.25  & 0.29 \\
                            {}  & 6.0 & 0.15 &  0.21\\
         \midrule[0.5pt]
         \multirow{2}{*}{0.02}  & 2.0 & 0.08  &  0.08 \\
                            {}  & 4.0 & 0.07 &  0.14\\
                            {} & 5.5 & 0.18  &  0.23\\
                            {}  & 6.0 & 0.09 &  0.16\\
         \midrule[0.5pt]
         \multirow{2}{*}{0.03}  & 2.0 & 0.08  &  0.09\\
                            {}  & 4.0 & 0.06 &  0.12\\
                            {} & 5.5 & 0.17  & 0.20\\
                            {}  & 6.0 & 0.08 &  0.12\\
         \bottomrule
    \end{tabular}
    \label{t:zhucomp}
\end{table}



\subsection{Generating Microstructures using Statistical Invariances}

 \begin{figure}[t]
 \centering
 \hspace{-25pt}\begin{minipage}{0.45\textwidth}
 \centering
		\setlength{\tabcolsep}{1.5pt}
		\renewcommand{\arraystretch}{0.8}
		\newcommand{\wah}{0.175}
		\begin{tabular}{ccccc}
            \includegraphics[width=\wah\linewidth ]{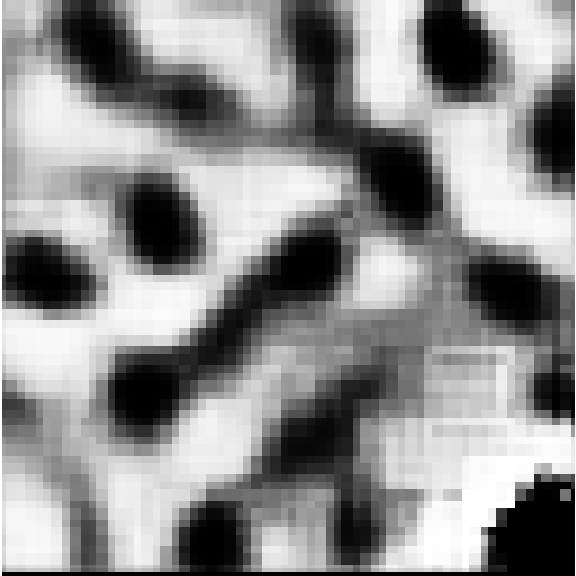} &
			\includegraphics[width=\wah\linewidth ]{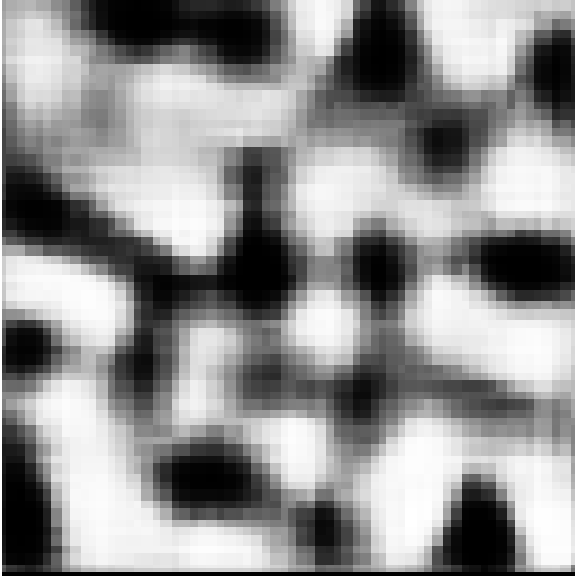} &
			\includegraphics[width=\wah\linewidth ]{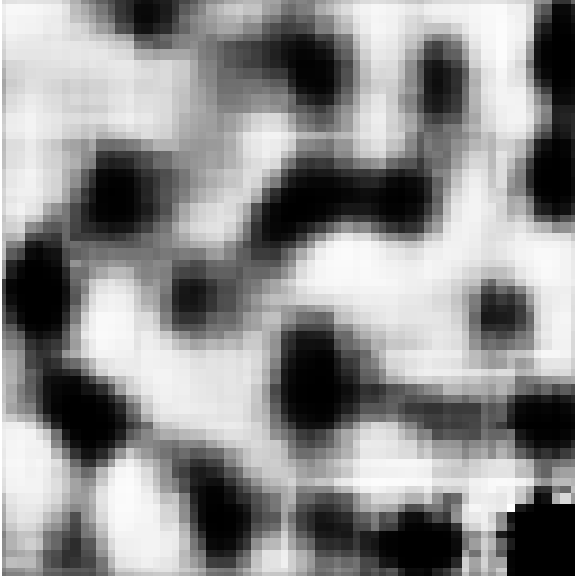} &
			\includegraphics[width=\wah\linewidth ]{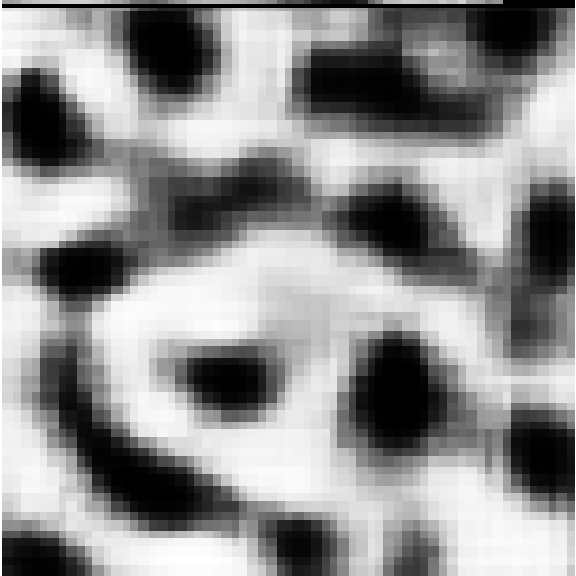} &
			\includegraphics[width=\wah\linewidth ]{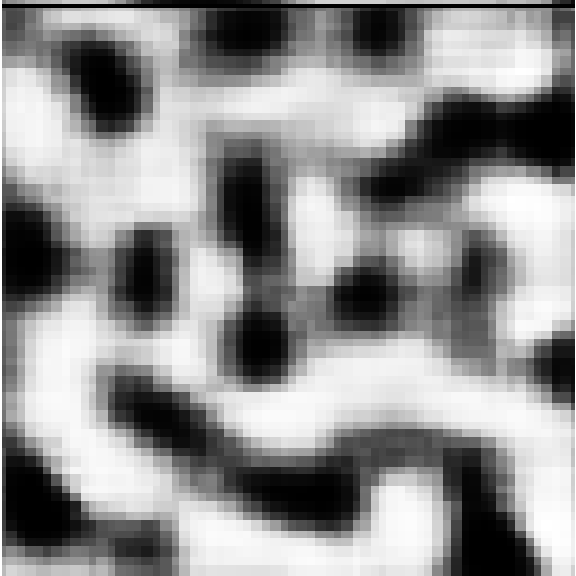} \\
			\includegraphics[width=\wah\linewidth ]{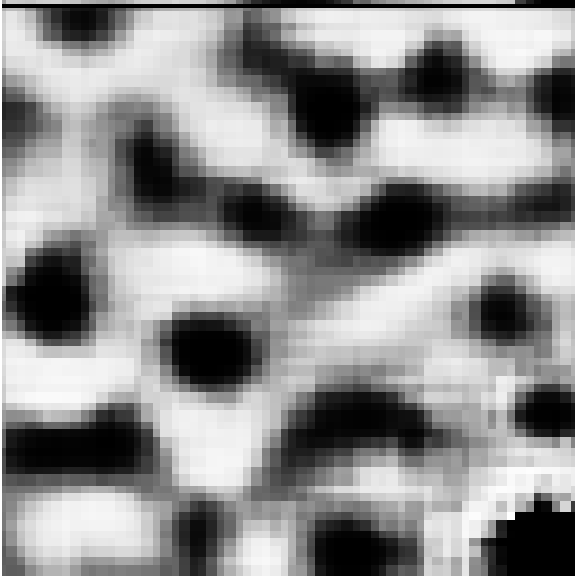} &
			\includegraphics[width=\wah\linewidth ]{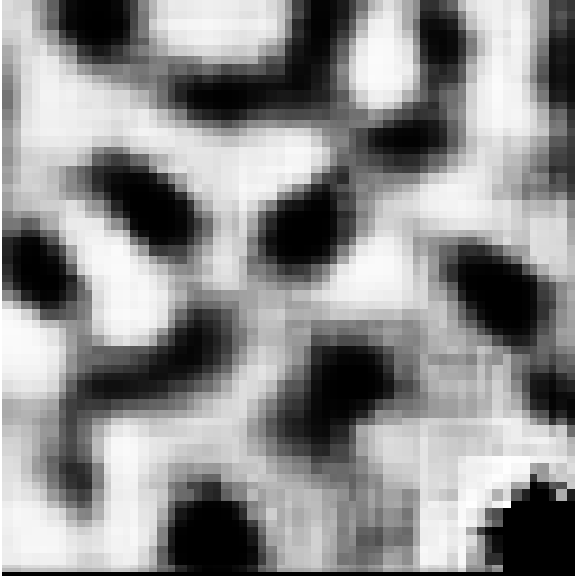} &
			\includegraphics[width=\wah\linewidth ]{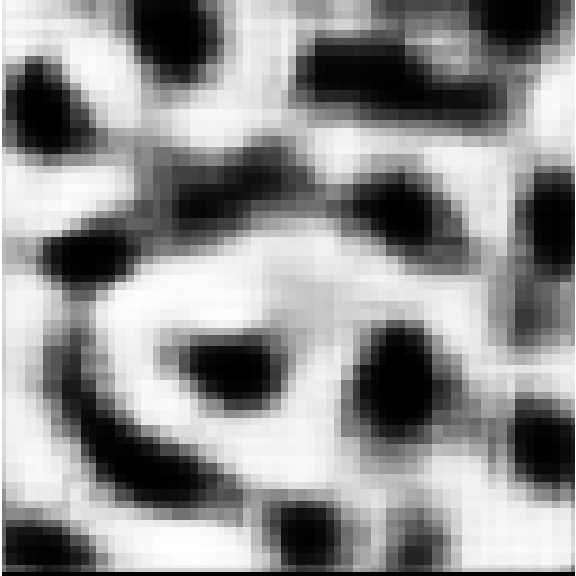} &
			\includegraphics[width=\wah\linewidth ]{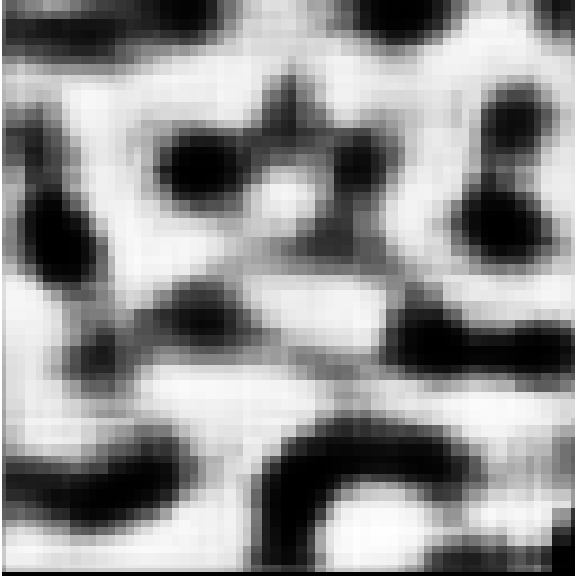} &
			\includegraphics[width=\wah\linewidth ]{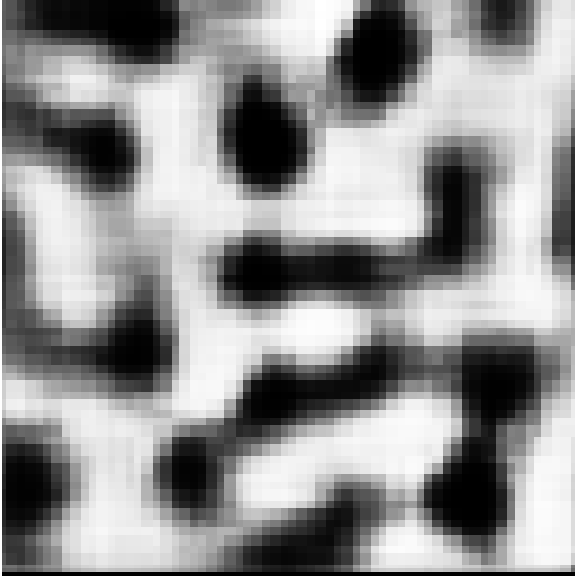}  \\ 
			\includegraphics[width=\wah\linewidth ]{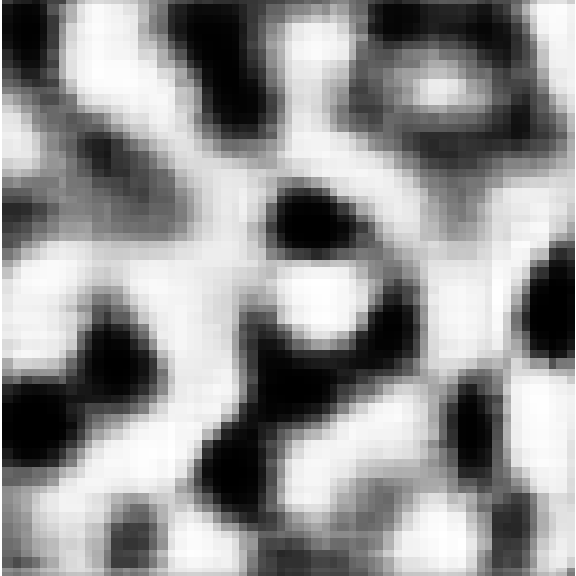} &
			\includegraphics[width=\wah\linewidth ]{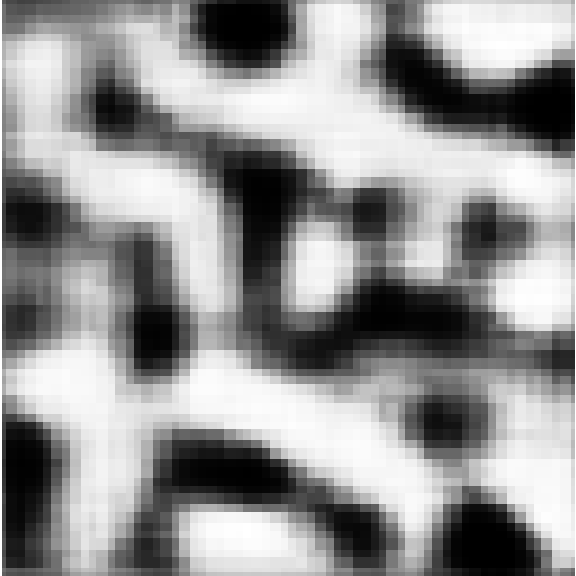} &
			\includegraphics[width=\wah\linewidth ]{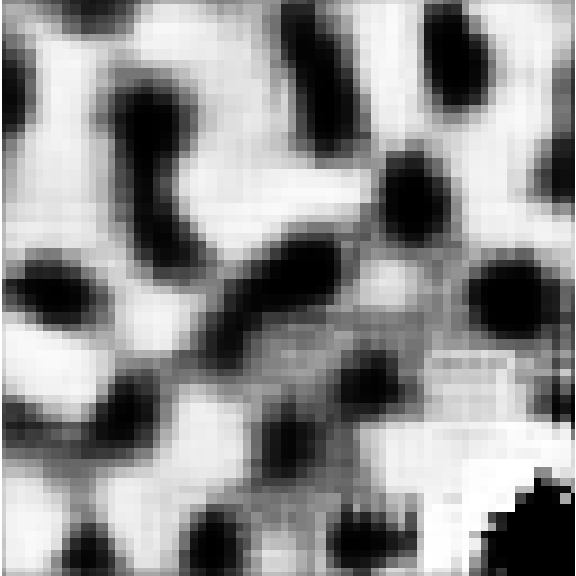} &
			\includegraphics[width=\wah\linewidth ]{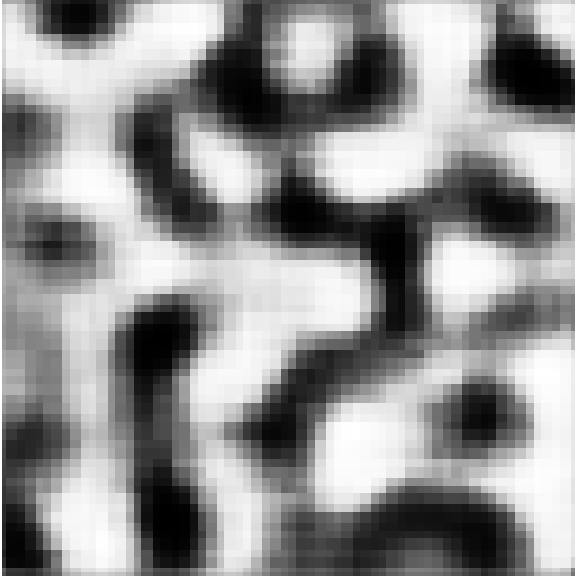} &
			\includegraphics[width=\wah\linewidth ]{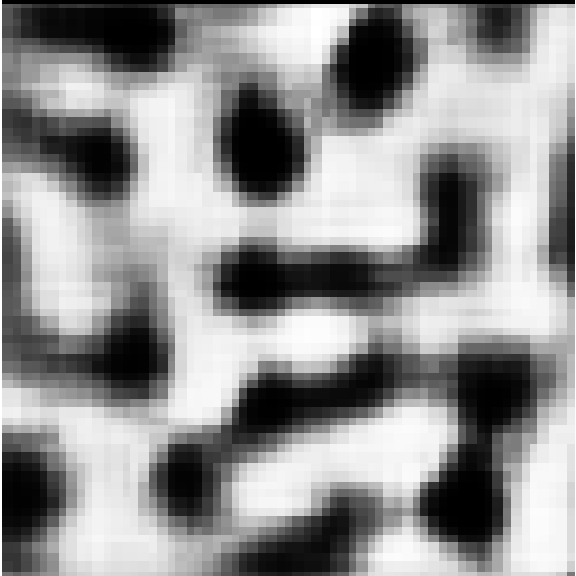}  \\ 
			\includegraphics[width=\wah\linewidth ]{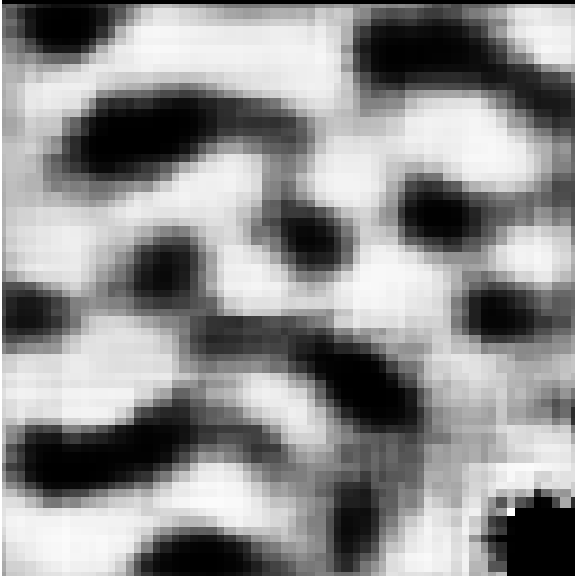} & 
			\includegraphics[width=\wah\linewidth ]{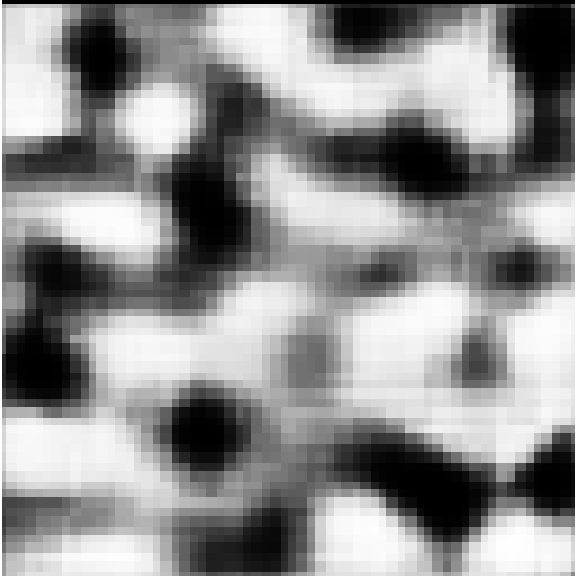} & 
			\includegraphics[width=\wah\linewidth ]{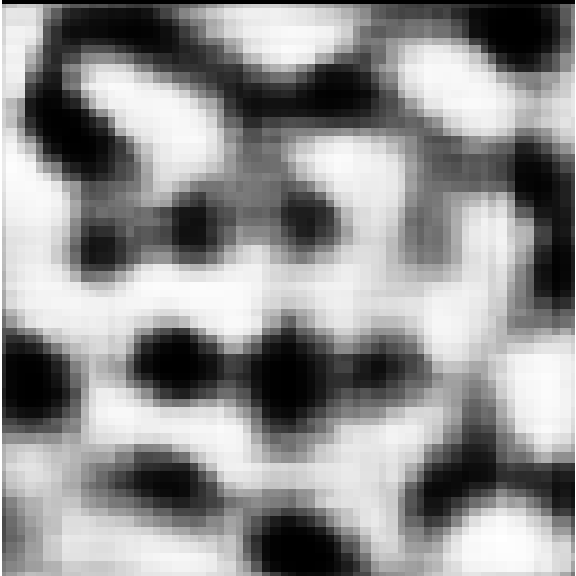} &
			\includegraphics[width=\wah\linewidth ]{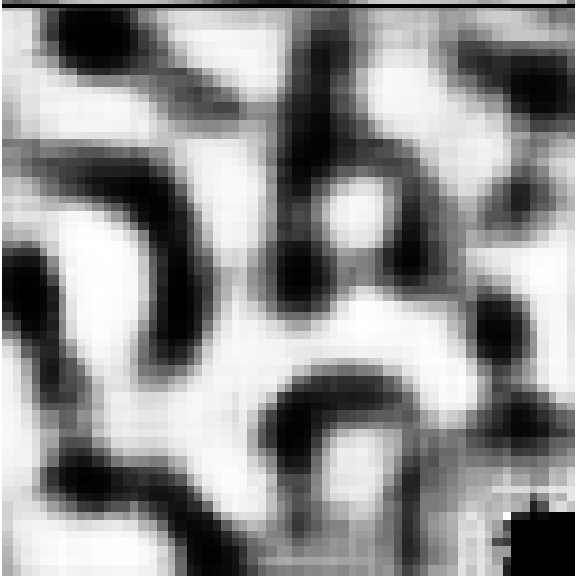} &
			\includegraphics[width=\wah\linewidth ]{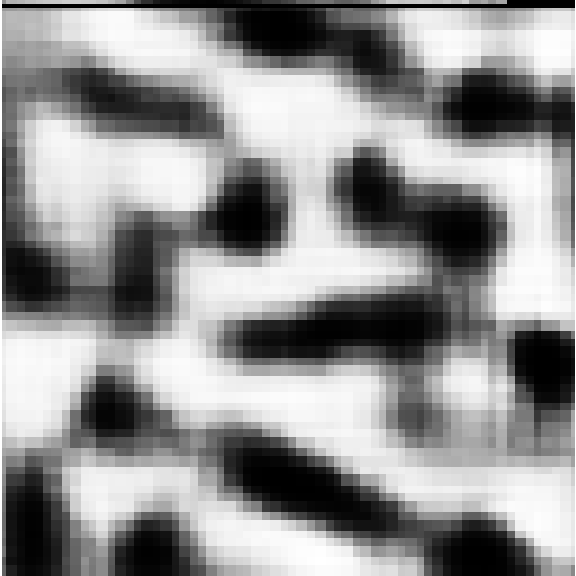}  \\	
		\end{tabular}
     \end{minipage}
 \hspace{10pt}
 \begin{minipage}{0.4\textwidth}
 	\renewcommand{\arraystretch}{1.2}
	\begin{threeparttable}[t]
		\begin{tabular}{ll} \toprule
			Model type & Time (s) \\ \midrule
			Numerical solution~\cite{Wodo2012} &  \\ \cmidrule{1-1}
			Generating $1$ microstructure$^\dagger$ & $1.84s$ \\
			Total time for $100000$ images$^\dagger$ & $\mathbf{184000s}$ \\ \hline
			\textbf{InvNet} (our approach) &  \\ \cmidrule{1-1}
			Training time$^\times$ & $57600s$ \\
			Generating $1$ microstructure$^\times$ & $0.0110s$ \\
			Total time for $100000$ images$^\times$ & $\mathbf{58700s}$ \\ \hline
		\end{tabular}
		\begin{tablenotes}\scriptsize
			\item[\scriptsize{$\dagger$}] Uses Intel 4-core CPU with 32 GB RAM.
			\item[\scriptsize{$\times$}]Uses 1 NVIDIA Tesla V100 GPU, 32 GB GDDR5 on TensorFlow GPU version 1.4.
		\end{tablenotes}
	\end{threeparttable}
 \end{minipage}
 \vspace{2.0mm}
\caption{\sl (Left) InvNet generated microstructures for fixed $1^{st}$ and $2^{nd}$ moments. (Right) Comparison with simulation times for generating $10^5$ microstructure images using numerical  methods~\cite{Wodo2012}. 
}
\label{fig:comparison}
\end{figure}

In computational material science, material distribution is often represented by an image describing the arrangement of constituents within a material. The statistical properties of such images govern the physical properties of the underlying material. Synthesizing microstructures adhering to specific statistical properties is, therefore, a crucial component of new material discovery. 

We focus on binary microstructures (corresponding to black/white images). 
Generally, the first and second moments of the image are useful statistical descriptions. Formally, we consider:
(i) the $1^{st}$ moment, $p_1$, also called the \emph{volume fraction}, and 
(ii) the $2^{nd}$ moment, $p_2$, also called the \emph{two-point correlation}. The former is a scalar, while the latter is a function. 
We focus on binary microstructures exhibiting \emph{phase separation}; their dynamics are governed via the well-known Cahn-Hilliard (CH) equation~\cite{cahn1958free}. This is a fourth-order nonlinear PDE, and its solution requires a significant amount of compute time (see Table~\ref{fig:comparison}).

We remedy this by training an InvNet to generate microstructures adhering to desired statistical properties. Since the first and second moments are differentiable functions, we encode the desired statistical properties into the InvNet formulation using the invariances:
\begin{equation}
    L_I = \lambda_1 \| f_{p_1}(G_\theta(\rvz)) - p_1^*\|_2^2 + \lambda_2 \| f_{p_2}(G_\theta(\rvz)) - \rvp_2^* \|_2^2
    \label{eq:p1_p2}
\end{equation}
where $f_{p_i}$ represent the functional forms of the moments, and $ p_1^*,\rvp_2^*$ are target values of the moments.

For training the InvNet, we use a publicly available dataset of 2D binary microstructures containing $\sim34k$ images across the wide range of statistical moments~\cite{pokuri_balaji_sesha_sarath_2019_2580293} (refer appendix~\ref{sec:app-micro} for details). 
The results in Fig.~\ref{fig:comparison} show the generated images adhering to target invariances along with the running time of our approach with existing numerical approaches. We note that the training time incurred by InvNet is amortized over the generation time required to simulate many microstructures; to generate 100,000 microstructures InvNet already obtains speedups over existing numerical solutions.

\section{InvNet: Theoretical Analysis}
\label{sec:theory}

We supplement our empirical results with rigorous theoretical analysis of InvNet training for some special (but illustrative) cases.

\paragraph{Equilibrium analysis.}
Two-player minimax games require that the generator and discriminator reach a (pure) Nash equilibrium. We analyse such equilibria for InvNet for two cases: (1) a standard WGAN discriminator, (2) a modified discriminator with the complementary projection operator. All theorems and proofs can be found in Appendix~\ref{sec:app-theory}.

We consider the motivating example of motif invariances for a simple data model similar to that described in Daskalakis~\etal~\cite{daskalakis2017training}. Let $\mathcal{D}$ be the data generated by a multivariate normal distribution, $\mathcal{N}(v, \id)$ where the mean, $v \in \sR^d$, needs to be learned. We stipulate a motif, $\rvm^*$, defined over a set of coordinates $\Omega$ that need to obey specific values. 

Similar to~\cite{daskalakis2017training}, we simplify the generator, $G_\theta(.)$ and the discriminator, $D_\psi(.)$ as follows:
\begin{align*}
D_{\psi}(\rvx) = \langle \psi, \rvx \rangle,~~G_{\theta}(\rvz) = \rvz + \theta \, .
\end{align*}
Following the InvNet model, we define the two-player game as an minimax optimization. The overall loss function becomes:
\begin{align}
   \bar{L}(\theta, \psi) &= \E_{\rvx \in \sP_\mathcal{D}} \left[D_\psi(G_\theta(\rvx))\right] - \E_z\left[(D_\psi(G_\theta(\rvz)))\right] + \mu \| \gP_{\Omega}(G_\theta(z)) - \rvm^* \|_2^2
    \label{eq:inv_net_lang}
\end{align}
Of key interest in such games is a (pure) Nash equilibrium point, where the gradients of the loss equate to $0$. The standard approach to achieve a Nash equilibrium of \Eqref{eq:inv_net_lang} is to train the parameters of both the networks ($\theta$ and $\psi$) via simultaneous gradient descent updates, i.e. 
\begin{align*}
\psi_{l+1}= \psi_{l} + \eta \nabla_{\psi}L(\theta_l, \psi_l),~~\theta_{l+1} = \theta_l - \eta\nabla_{\theta}L(\theta_l, \psi_l) \, .
\end{align*}
If we consider the limiting case where the true expectations are optimized (i.e., we have infinite samples), one can simplify the InvNet loss into the form:
\begin{align*}
\bar{L}(\theta, \psi) 
                     &= \langle \psi, v-\theta \rangle + \frac{\mu}{2}\| \gP_{\Omega}(\theta) - \rvm^*\|^2 .
\end{align*}
Taking gradients and setting to zero, we get the (unique) equilibrium point at $\theta = v~, \psi = \mu(\gP_{\Omega}(\theta - \rvm^*))$\footnote{Note that the same approach for a standard WGAN setup (without the invariance term), the equilibrium point is simply $(\theta^*, \psi^*)= (v, 0)$.}.
Surprisingly for InvNet, the presence of the motif invariance term fails to be reflected in the generator, and is entirely compensated by the discriminator. This is fine if the mean of the data, $v$, \emph{indeed} exhibits the motif (i.e., $P_\Omega(v) = \rvm^*$), but if not, then there is no way of incorporating it into the generated data.

To alleviate the issue, we modify the discriminator to ignore the effect of the motif invariance; we force the discriminator to only consider the complementary subset of coordinates, $\gP_{\Omega^c}(\rvx)$.
The modified discriminator ($D_{\psi}(\gP_{\Omega^c}(\cdot))$) becomes:
\begin{align}
   \bar{L}(\theta, \psi)= \langle \gP_{\Omega^c}\psi, \gP_{\Omega^c}(v-\theta) \rangle + \frac{\mu}{2}\|\gP_{\Omega}(\theta - \rvm^*)\|_2^2 .
   \label{eq:inv_mod_disc}
\end{align}
Computing gradients and setting to zero, we find that the new (unique) Nash equilibrium is given:
\begin{align*}
\theta &= \begin{bmatrix} 
  \gP_{\Omega} m^*\\ 
\gP_{\Omega^c}v
\end{bmatrix},~ \psi = \begin{bmatrix} 
  \gP_{\Omega} \kappa\\ 
0
\end{bmatrix};
\textnormal{with $\kappa$ as an arbitrary constant vector.}
\end{align*}
Remarkably, in this case, the generator has learned to reproduce the motif $\rvm^*$ in the coordinates specified by $\Omega$, and to copy the data mean onto the rest of the coordinates. This suggests that under the modified discriminator, our model enforces better invariances. Recall that we empirically observed similar trends in our experiments (Fig.~\ref{fig:motif}) with motifs planted on MNIST images; such a modification in the discriminator resulted in better discriminator loss profile while generating samples with desired motifs.


\paragraph{Training Dynamics.} While the above analysis is with respect to InvNet equilibrium points, we need a computationally efficient means to achieve such equilibria. For this, we consider the standard gradient descent (GD) algorithm.
The linear system of GD iterates for InvNet as defined in \Eqref{eq:inv_net_lang} is described as,
\begin{align*}
\psi_{l+1} = \psi_{l} + \eta \nabla_{\psi}L(\theta_l, \psi_l),~~\theta_{l+1} = \theta_l - \eta\nabla_{\theta}L(\theta_l, \psi_l) .
\end{align*}
We show that, for any value of \(\eta\), however small, the system diverges resulting in unstable dynamics. The same holds true for the game defined in \eqref{eq:inv_mod_disc}. The proof (in Appendix~\ref{sec:app-theory}) involves analyzing the singular values of the dynamical system defined by the gradient updates, but they turn out to be always greater than $1$. 
A standard approach to repair such unstable systems is to employ momentum-based adjustments. Following the lead of \cite{daskalakis2017training, liang2018interaction, mokhtari2019unified}, we propose the extra-gradient descent (EGD) method for stabilizing the training of InvNet. In  EGD, the key idea is to modify the update direction by taking the gradient at an interpolated point between the current and the next point. Specifically, given a stepsize $\eta > 0$, the EGD update for Eqn.~\ref{eq:ganloss} consists of two steps:
\begin{align}
    \label{eq:eg-half}
    \theta_{l+\frac{1}{2}} = \theta_{l} -\eta\nabla_{\theta}L(\theta_{l}, \psi_{l}), & ~~\psi_{l+\frac{1}{2}} = \psi_{l} + \eta\nabla_{\psi}L(\theta_{l}, \psi_{l}) \\ \nonumber
    \theta_{l+1} = \theta_{l} -\eta\nabla_{\theta}L(\theta_{l+\frac{1}{2}}, \psi_{l+\frac{1}{2}}),& ~~\psi_{l+1} = \psi_{l} + \eta\nabla_{\psi}L(\theta_{l+\frac{1}{2}}, \psi_{l+\frac{1}{2}}) . \nonumber
\end{align}
Repeating the analysis as before, we can prove that the iterates form a contractive mapping for an informed choice of $\eta$. Further, the result also extends to the modified discriminator case. 

\section{Discussion and Conclusions}
\label{sec:disc}

We have proposed InvNets, a natural extension of GANs that enables specifying additional structural/statistical invariances in generated samples. 
Our approach relies on representing the invariance as an additional loss term that we alternately optimize in addition to the standard adversarial training in GANs. We also show a number of stylized and real-world applications. 

Several potential directions of research remain. In the context of solving PDEs, limiting the generator to images restricts its use to problems in two dimensions. A compelling extension would be to extend InvNets for other domains such as graphs. 

Our theoretical analysis above provides unexpected insights. If there is a mismatch between the data and the desired invariance, the discriminator in InvNet will necessarily need to be modified. However, if the invariance is part of the true data distribution, the discriminator learns to incorporate it; cf.\ our experiment of generating binary microstructures with required statistical properties. 

While stylized, our theory points to several open questions about adversarial models and the interaction of multi-task objectives. An interesting question that arises is that of designing discriminators to ignore nonlinear invariances and thereby allow for training GANs with incomplete data. Finally, a more rigorous analysis into the dynamics of non-limiting cases of training InvNet is warranted.

\section*{Acknowledgements}

This work was supported by the DARPA AIE program under grant DARPA-PA-18-02-02, the National Science Foundation under grants NSF CCF-1750920, NSF DMREF 1435587, U.S. AFOSR YIP grant FA9550-17-1-0220, GPU gift grants from NVIDIA Corporation, and a faculty fellowship by the Black and Veatch Foundation.


{
\bibliographystyle{unsrt}
\bibliography{neurips_bib}
}
\clearpage
\appendix
\normalsize
\appendix
\section*{Appendix}
\textit{\textbf{Notation.}} We represent vectors in lowercase boldface, $\rvv$ whereas matrices are uppercase boldface, $\rmV$. Given an image, $\rmU$, we represent the spatial derivative with respect to axis, $x$ as $\rmU_x$. Additionally, the gradient of a scalar, $L$ with respect to a vector, $\rva$ is represented as $\nabla_a L$. $\id$ represents identity matrix.
\section{Theorems and Proofs}
\label{sec:app-theory}
\subsection{Instability of Gradient Descent}
\begin{theorem}\emph{(Instability of GD for InvNet Training)}. Consider InvNet minimax game given by Eqn.~\ref{eq:inv_net_lang} with a linear discriminator $D_{\psi}$, an additive displacement generator $G_{\theta}$ and motif $\rvm^*$. Then the GD dynamics, 
	\begin{align*}
	\psi_{l+1} &= \psi_{l} + \eta \nabla_{\psi}L(\theta_l, \psi_l) \\
	\theta_{l+1} &= \theta_l - \eta\nabla_{\theta}L(\theta_l, \psi_l)
	\end{align*}
	with the learning rate $\eta$ leads to unstable dynamical system for both (i) standard discriminator, and (ii) modified discriminator.
\end{theorem}
\proof{Applying GD update rules on Eqn.~\ref{eq:inv_net_lang}, the dynamical system for the standard generator case becomes:
	\begin{align*}
	\theta_{l+1} &= \theta_l + \eta \psi_{l}- \eta \mu\gP_{\Omega}(\theta_l - \rvm^*) \\
	\psi_{l+1} &= \psi_{l} + \eta(v-\theta_l) \\
	\implies 
	\begin{bmatrix}
	\theta_{l+1} \\
	\psi_{l+1}
	\end{bmatrix} &=
	\underbrace{
		\begin{bmatrix}
		\id-\eta\mu\gP_{\Omega} & \eta \id \\
		-\eta \id & \id
		\end{bmatrix}}_\text{$\rmA$}
	\begin{bmatrix}
	\theta_{l} \\
	\psi_{l}
	\end{bmatrix} + 
	\begin{bmatrix}
	\eta\mu\gP_{\Omega}\rvm^* \\
	\eta v
	\end{bmatrix}
	\end{align*}
	We study the singular values of the matrix $\rmA$ to characterize the stability of the training dynamics. 
	\begin{align*}
	    \rmA\rmA^T 
	           &=\begin{bmatrix}
	                \id-\eta\mu\gP_\Omega -\eta\mu\gP_{\Omega}^T +\eta^2\mu\gP_\Omega\gP_\Omega^T & \eta^2\mu\gP_\Omega\gP_\Omega \\
	                \eta^2\mu\gP_\Omega^T & (\eta^2 + 1)\id
	           \end{bmatrix} \\
	           &= \begin{bmatrix}
	                    \id & 0 \\
	                    0 & \id
	               \end{bmatrix} + 
	               \begin{bmatrix}
	                    -\eta\mu\gP_\Omega -\eta\mu\gP_\Omega^T +\eta^2\mu\gP_\Omega\gP_\Omega & \eta^2\mu\gP_\Omega\gP_\Omega^T \\
	                    \eta^2\mu\gP_\Omega^T & \eta^2\id
	                \end{bmatrix}
	\end{align*}
	
	Using Weyl's Inequality, we show that all singular values of $\rmA$ are lower bounded by $1$. This implies that the linear system of equations is unstable. 
	
	Similarly, for the modified discriminator case, the dynamical system becomes:
	
	\begin{align*}
	\begin{bmatrix}
	\theta_{l+1} \\
	\psi_{l+1}
	\end{bmatrix} &=
	\underbrace{
		\begin{bmatrix}
		\id -\eta\mu\gP_{\Omega} & \eta\mu\gP_{\Omega^c}  \\
		-\eta \id \gP_{\Omega^c} & \id
		\end{bmatrix}}_\text{A}
	\begin{bmatrix}
	\theta_{l} \\
	\psi_{l}
	\end{bmatrix} + 
	\begin{bmatrix}
	\eta\mu\gP_{\Omega}\rvm^* \\
	\eta \gP_{\Omega^c}v
	\end{bmatrix}
	\end{align*}
	Again, in this case too the singular values are lower bounded by $1$ for any choice of positive $\eta$ and $\mu$. Hence, for both the standard and modified discriminators the training dynamics are unstable for GD updates. \qedsymbol
}
\subsection{Analysis of Extra Gradient Training Dynamics}
\begin{theorem}(\emph{Stability of EGD for InvNet Training}). Consider InvNet minimax game given by Eqn.~\ref{eq:inv_net_lang} with a linear discriminator $D_{\theta}$, an additive displacement generator $G_{\theta}$ and motif $\rvm^*$. Then the EGD with the learning rate $\eta$ leads to stable dynamical system for standard discriminator.
\end{theorem}
\proof{
	Applying EGD updates on our original InvNet formulation:
	\begin{align*}
	\theta_{l+\frac{1}{2}} & = \theta_{l} -\eta\left[\psi_{l} - \mu\gP_{\Omega}(\theta_l - \rvm^*)\right] \\ 
	\psi_{l+\frac{1}{2}} & = \psi_{l} + \eta\left[(v-\theta_{l})\right] \\
	\theta_{l+1} & = \theta_{l} + \eta\left[\psi_{l} + \eta\left((v-\theta_{l})\right) - \mu\gP_{\Omega}(\theta_{l} -\eta\left(\psi_{l} - \mu\gP_{\Omega}(\theta_l - \rvm^*)\right) - \rvm^*)\right] \\
	& = \left[\id-\eta^2 -\mu\gP_{\Omega} + \eta\mu^2\gP_{\Omega}\right]\theta_l + (\eta-\eta\mu\gP_{\Omega})\psi_l + \eta^2v - \eta\mu^2\gP_{\Omega}\rvm^* \\
	\end{align*}
	Similarly,
	
	\begin{align*}
	\psi_{l+1} & = \psi_{l} + \eta\left(v-\theta_k-\eta\psi_k + \eta\mu\gP_{\Omega}(\theta_l - \rvm^*)\right) \\
	& = \left(-\eta + \eta\mu\gP_{\Omega}\right)\theta_l + (1-\eta^2)\psi_l + \eta v
	\end{align*}
	
	For stability analysis:
	
	\begin{align*}
	\begin{bmatrix}
	\theta_{l+1} \\
	\psi_{l+1}
	\end{bmatrix} &=
	\underbrace{
		\begin{bmatrix}
		\id-\eta^2 -\mu\gP_{\Omega} + \eta\mu^2\gP_{\Omega} & \eta - \eta\mu\gP_{\Omega} \\
		-\eta + \eta\mu\gP_{\Omega} & \id-\eta^2
		\end{bmatrix}}_\text{$\rmA$}
	\begin{bmatrix}
	\theta_{l} \\
	\psi_{l}
	\end{bmatrix} + 
	\begin{bmatrix}
	\eta^2v - \eta\mu^2\gP_{\Omega}\rvm^* \\
	\eta^2 v
	\end{bmatrix}
	\end{align*}
	
	Here, analysis of singular values of matrix $A$ will help us determining the stability of our method. Matrix $A$ can be written as,
	\begin{align*}
	\underbrace{
		\begin{bmatrix}
		\id(1-\eta^2) & \eta \id \\
		-\eta \id  & \id(1-\eta^2)
		\end{bmatrix}}_\text{$\rmB$} - \eta\mu
	\underbrace{
		\begin{bmatrix}
		(1-\mu)\gP_{\Omega}  & \gP_{\Omega}\\
		- \gP_{\Omega} & 0
		\end{bmatrix}}_\text{$\rmC$}
	\end{align*}
	
	All singular values of $\rmB$ are of the form:
	\[
	\sqrt{(1-\eta^2)^2 + \eta^2}
	\]
	Also, for singular values of $\rmC$ is less than or equal to $c\eta\mu$, with some constant $c$. We use Weyl's inequality for matrices $\rmB$ and $\rmC$
	where $A=B+C$. For careful choice of $\eta$,
	\[
	\sigma_{\rmA} \leq \sigma_{\rmB} + \sigma_{\rmC} =  \sqrt{(1-\eta^2)^2 + \eta^2} + c\eta\mu \leq 1
	\]
	Which proves the stability of EGD updates for our case. \qedsymbol
}

\section{Generating PDE solutions with InvNet: Additional Details}
\label{sec:supp_pde}

We primarily motivate InvNet for generating solutions to complex physical systems. The dynamics of such systems are often described partially/completely by partial differential equations. Recent literature~\cite{raissi2017physics1, raissi2017physics2, zhu2019physics, zhu2018bayesian} create surrogate to PDEs with deep generative neural networks. However, this involves training a new neural network for every new instance of the equation. We, therefore propose a variant of InvNet to generate solutions to a family of PDEs for a large class of boundary conditions. 

Considering the family of PDEs as an optimal discriminator (parameterized over the coefficients), we additionally impose a structural invariance in the form of boundary conditions. Note that given a particular tuple of initial condition and coefficients, the solution to the PDE is unique (given a solution exists). We modify the input vector accordingly to be a representation of the initial conditions and the coefficient. Consequently, we drop the stochastic part of our input. The optimality of the discriminator leads to the data dependent discriminator loss going to $0$ as all data is invariably going to satsify the said family of equations. This negates the need for training data as the InvNet can completely model the class of these PDEs simply by being conditioned on the coefficients and boundary conditions.

We present an example of solving a class of nonlinear PDEs for a set of boundary conditions using InvNet. As a demonstration, we solve the well known Burgers' equation~\cite{burgers} which is a nonlinear PDE defining the advection of a wave through a diffusive liquid. We provide a brief introduction and description of the class of these equations.

\textbf{Burgers' Equation. } Burgers' equation is a fundamental non linear PDE that describes the advective flow of a wave in a fluid. The viscid form as described in \Eqref{suppeq:viscid} analyses the dynamics of flow in a diffusive liquid where energy dissipates as the wave travels.

\begin{equation}
    \rmU_{,t} + \rmU\rmU_{,x} = \nu \rmU_{,xx}
    \label{suppeq:viscid}
\end{equation}

A simpler case is that of an non-diffusive ideal liquid (i.e. $\nu = 0$). However, the \emph{inviscid} Burgers' equation describes a energy conserving system that is prone to shocks.

The solution for the inviscid Burger's equation is of the form:
\begin{equation*}
     u(x,t) = f(x-ut)
\end{equation*}
where $f(\cdot)$ is the function that defines the initial condition, $u(x,0)$.

Representing the solution, $\rmU$ as an image where the axes correspond to time and space, we train an InvNet to surrogate the PDE.

\textbf{Implementation details.} We use the invicid form of the Burgers' equation to construct a residual function of the form,
\begin{equation}
    \| \rmU_{,t} + \rmU\rmU_{,x} \|_F^2
    \label{suppeq:pde_res}
\end{equation}

The spatial derivatives in this case can then be approximated using image gradients, similar to the method used in \cite{zhu2019physics}. We use Sobel directional derivative operators~\cite{sobel19683x3} to approximate the two derivative terms. Since these are linear operations, gradients with respect to the loss function can easily be calculated using back-propogation. However, there are a few conditions that need to be considered for numerical analysis of PDEs.

The first one is the Courant–Friedrichs–Lewy (CFL) condition that requires that the time step be a fraction of the spatial step for convergence to a solution. 
We enforce this by simply discretizing $t$ and $x$ with different step-sizes. This can be achieved by normalizing the corresponding sobel operator with two separate constants. For $t$, we use $5/64$ whereas for $x$, we use $1/64$. This results in a domain $x \in [0,1]$ and $y \in [0, 0.2]$ and a Courant number of $0.2$, which ensures convergence.
This is surprising since the InvNet predicts the full space-time solution, rather than a time marching scheme where the CFL satisfaction ensure a numerical solution. This interesting behavior warrants additional investigation.

Additional considerations to avoid border effects include replicating the boundaries as well as modifying the Sobel operator at the boundaries to emulate forward differences. We follow a similar methodology as described in \cite{zhu2019physics} to correct spurious boundary effects.

The second term for InvNet is the structural invariance, defined here as the initial condition. For the initial condition, we consider a family of functions of the form, 
\begin{equation*}
    \frac{1}{2}(1 - \cos (2\pi c x))
\end{equation*}
where $c \sim \text{Unif}(3.0, 6.0)$. Here, we are limited to sample $c$ to a maximum of $6$ to avoid aliasing effects as the generator has output dimensions of $64\times64$. However, our choice of the output dimension is simply due to computational constraints as GANs have been shown to generate large high definition images. 

For the viscid case, the second derivative is similarly approximated using a Gaussian smoothed Laplacian operator. An additional consideration for the viscid case is that the PDE loss is now parameterized on a scalar viscosity term, $\nu$. In physical terms, $\nu$ controls the rate of energy dissipation of the wave as it travels through the diffusive fluid. For InvNet, we sample $\nu$ from a distribution, which allows for the generator to effectively represent a solution space for a family of Burger's viscid equations. Additionally conditioning the generator on $\nu$ allows for sampling solutions in a controlled fashion for a specific PDE.

The alternating optimization approach here is essential to ensure adversarial interaction between the PDE loss, that minimizes all points on $\rmU$ at once and the invariance loss that only considers a single column. We train two instances of InvNet to solve the inviscid and the viscid forms respectively. \Figref{fig:burgerssoln} shows examples generated by the two models. We especially draw attention to the generalization capabilities of our model as evidenced by column (1) in the figure, where the generator solves for an unseen initial condition ($c=2.0$).

\section{Microstructures Generation using InvNet: Additional Details}
\label{sec:app-micro}
\subsection{Motivation}
An overarching theme of materials research is the design of material distributions (also called microstructure) so that the ensuing material exhibits tailored properties. In microstructure-sensitive design, quantifying the effect of microstructure features on performance is critical for the efficient design of application-tailored devices. Microstructures are represented as binary images indicating the arrangement of constituent materials within the mixture. The statistical properties of such microstructural images are useful in predicting the physical and chemical properties of the mixture material - thus aiding into faster material discovery. To obtain a material with desired property, a microstructure having the corresponding statistical property need to be generated. We feed in such statistical properties as the invariances in our framework, and come up with a generative model that can sample from the set of all microstructures adhering to desired statistical properties. We propose both a data-driven and data-free generative network for synthetic microstructures adhering the invariances for the training.
\subsection{Preliminaries}
In the context of microstructure generation problem, we consider the underlying material to be a two-phase homogeneous, isotropic material. 
Our setup for statistical characterization of microstructure follows with Torquato \etal~\cite{torquato2013random}. Consider an instance of the two-phase homogeneous isotropic material within $d$-dimensional Euclidean space $\mathbb{R}^d$ (where $d \in \{2,3\}$). A phase function $\phi(\cdot)$ is used to characterize this two-phase system, defined as:
\begin{equation}
    \phi^{(1)}(\mathbf{r}) = \begin{cases}
                                1, \mathbf{r} \in V_1,\\
                                0, \mathbf{r} \in V_2,
                        \end{cases}
\end{equation}
where $V_1 \in \mathbb{R}^d$ is the region occupied by phase 1 and $V_2 \in \mathbb{R}^d$ is the region occupied by phase 2. 

Given this microstructure defined by the phase function, $\phi$, statistical characteristics can be evaluated. These include the $n-$moments, ($n$-point correlation functions) for $n ={1,2,3,...}$. For homogeneous and isotropic media, These depend neither on the absolute positions of $n-$points, nor on the rotation of these spatial co-ordinates; instead, they depend only on relative displacements. The $1^{st}$-moment, $p_1$, commonly known as \emph{volume fraction}, is constant throughout the material. The volume fraction of phase 1, $p_1^{(1)}$, is defined as:
$$
p_1^{(1)} = \mathbb{E}_{\mathbf{r}} \phi^{(1)}(\mathbf{r}).
$$
The $2^{nd}-$moment is a function of $r$ and is defined as: 
$$
\rvp_2^{(1)}(r_{12}) = \mathbb{E}_{\mathbf{r_1},\mathbf{r_2}} \left[\phi^{(1)}(\mathbf{r}_1)\phi^{(1)}(\mathbf{r}_2)\right].
$$

The $2^{nd}$ moment (known as $2-$point correlation as well) is one of the most important statistical descriptors of microstructures. An alternate interpretation of $2^{nd}$ moment is the probability that two randomly chosen points $\mathbf{r}_1$ and $\mathbf{r}_2$ a certain distance apart both share the same phase. 

Henceforth we omit the superscript representing the phase and subscripts representing the spatial points for simplicity, and refer to volume fraction as $p_1$, and $2$-point correlation as $\rvp_2$. It can be shown that $\rvp_2(r = 0) = p_1$ and $\lim_{r\to\infty} \rvp_2(r) = p_1^2$.



In the training step, we use the above statistical properties as invariances for training the InvNet. The invariance loss $L_{I}(\cdot)$ can be defined as $l_2-$ loss:
\begin{equation}
    L_I = \lambda_1 \| f_{p_1}(G_\theta(\rvz)) - p_1^*\|_2^2 + \lambda_2 \| f_{\rvp_2}(G_\theta(\rvz)) - \rvp_2^* \|_2^2
    \label{eq:p1_p2_app}
\end{equation}
where $f_{p_i}$ represent the functional forms of the moments; $ p_1^*,  \rvp_2^*$ are target values of the moments.

We use the Binary 2D microstructures dataset~\cite{pokuri_balaji_sesha_sarath_2019_2580293} based on Cahn-Hilliard equation~\cite{cahn1958free} for training and testing. The dataset contains $\sim34k$ binary microstructures of size $101 \times 101$ obtained by sampling the evolving solutions across time. The dataset contains images with diverse values of $1^{st}$ and $2^{nd}$ moments, and implicitly exhibit higher moments too. For training, we resize the images to $64 \times 64$.

In Fig.~\ref{fig:hybridgraph}, we depict images generated by our InvNet, along with the target image from the real data that is used for setting the target values $p_1^*$ and $\rvp_2^*$. We also plot the distributions of $p_1$ and $\rvp_2$ of generated images, and show that they closely match with targets $p_1^*$ and $\rvp_2^*$. 


\subsection{Data-free InvNet for Microstructure Generation}
Further, we consider a case where set of our target images is completely defined by the statistical properties, i.e. there are no implicit features to learn. Only explicit structures to adhere to are $1^{st}$ and $2^{nd}$ moments, that should match given target values $p_1^*$ and $\rvp_2^*$ respectively. We obtain the $p_1^*$ and $\rvp_2^*$ by calculating moments on a real target image obtained by other numerical methods. In absence of any implicit features to learn, the discriminator in Eqn.~\ref{eq:inv_net} is no longer required, thus both the generator loss and discriminator loss vanish to left us with optimizing only over the invariance loss given by Eqn.~\ref{eq:p1_p2_app}. This exemplifies the simplest case of our model along with establishing the sufficiency of the invariance loss in training the generator. We show generated samples along with their distribution of $\rvp_2$ in Fig.~\ref{fig:invar1} and Fig.~\ref{fig:invar2a}. We observe that the invariances match quite well. Moreover, by setting the weights for elements of $\rvp_2$ in the loss function, we can make a part of $\rvp_2$ curve dominent in training, forcing that part to get enforced in a stricter manner, as depicted in Fig.~\ref{fig:invar2a}(c).  

\pgfkeys{/pgf/number format/.cd,1000 sep={\,}}
\usepgfplotslibrary{fillbetween}
 \begin{figure}[t]
 \begin{subfigure}[b]{0.32\textwidth}
 \centering
 \resizebox{\linewidth}{!}{
		\setlength{\tabcolsep}{0.2pt}
		\renewcommand{\arraystretch}{0.1}
		\begin{tabular}{ccccc}
				\multicolumn{2}{c}{\multirow{0}{0.130\linewidth}{\includegraphics[width=\linewidth,cframe=ube 0.4pt]{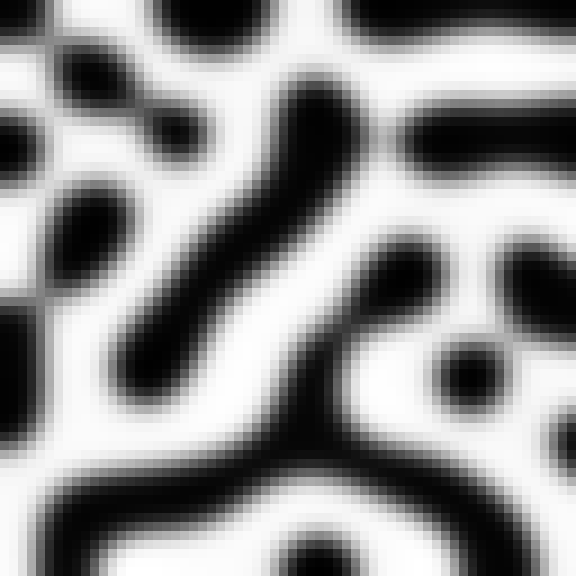}}} &
				   	& &   \\
					& &
			\includegraphics[width=\wa\linewidth ]{./fig/hybrid/img1.png} &
			\includegraphics[width=\wa\linewidth ]{./fig/hybrid/img3.png} &
			\includegraphics[width=\wa\linewidth ]{./fig/hybrid/img5.png}  \\
			 		& & 
			\includegraphics[width=\wa\linewidth ]{./fig/hybrid/img12.png} &
			\includegraphics[width=\wa\linewidth ]{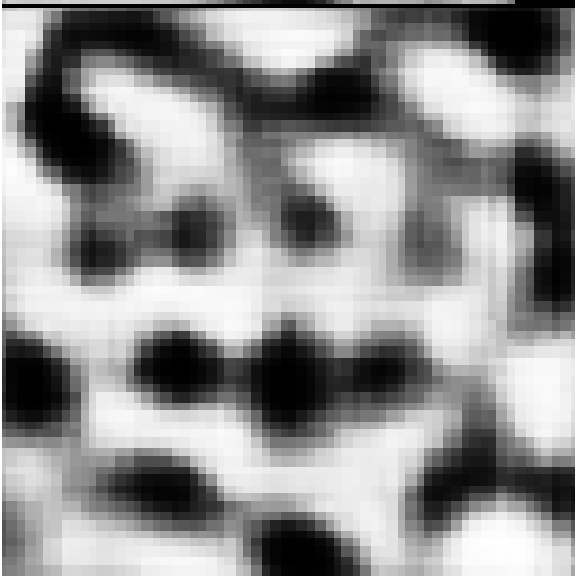} &
			\includegraphics[width=\wa\linewidth ]{./fig/hybrid/img14.png} \\	
			\includegraphics[width=\wa\linewidth ]{./fig/hybrid/img15.png} &
			\includegraphics[width=\wa\linewidth ]{./fig/hybrid/img16.png} &
			\includegraphics[width=\wa\linewidth ]{./fig/hybrid/img17.png} &
			\includegraphics[width=\wa\linewidth ]{./fig/hybrid/img18.png} &
			\includegraphics[width=\wa\linewidth ]{./fig/hybrid/img19.png}  \\ 
			\includegraphics[width=\wa\linewidth ]{./fig/hybrid/img20.png} &
			\includegraphics[width=\wa\linewidth ]{./fig/hybrid/img21.png} &
			\includegraphics[width=\wa\linewidth ]{./fig/hybrid/img22.png} &
			\includegraphics[width=\wa\linewidth ]{./fig/hybrid/img23.png} &
			\includegraphics[width=\wa\linewidth ]{./fig/hybrid/img24.png}  \\ 
			\includegraphics[width=\wa\linewidth ]{./fig/hybrid/img25.png} & 
			\includegraphics[width=\wa\linewidth ]{./fig/hybrid/img26.png} & 
			\includegraphics[width=\wa\linewidth ]{./fig/hybrid/img27.png} &
			\includegraphics[width=\wa\linewidth ]{./fig/hybrid/img28.png} &
			\includegraphics[width=\wa\linewidth ]{./fig/hybrid/img29.png}  \\	
		\end{tabular}}  
		\caption{\small{Generated images (target image on top left)}}
 \end{subfigure}
     \begin{subfigure}[b]{0.305\textwidth}
          \centering
          \resizebox{\linewidth}{!}{
          \begin{tikzpicture}
          \begin{axis}[axis background/.style={fill=seabornback, fill opacity=1},
        ylabel=\Large{$p_2$ correlation},
        grid style={line width=.1pt, draw=white},
        major grid style={line width=.1pt,draw=white},
        minor tick num=1,
        grid=both,
        xlabel= \Large{Radial distance},
        legend cell align=left,
        legend style={at={(1,1.25)}}]
\addplot[ultra thick,color=ube,x filter/.code={\pgfmathparse{\pgfmathresult*0.25}\pgfmathresult}] table[x=index, y=real, col sep=comma]{data/hybrid_p2.csv};
\addplot[thick,name path=max,color=amber,x filter/.code={\pgfmathparse{\pgfmathresult*0.25}\pgfmathresult}] table[x=index, y=max, col sep=comma]{data/hybrid_p2.csv};
\addplot[thick,name path = min, color=amber,x filter/.code={\pgfmathparse{\pgfmathresult*0.25}\pgfmathresult}] table[x=index, y=min, col sep=comma]{data/hybrid_p2.csv};
\addplot[color=amber, fill opacity=0.3] fill between[
    of = max and min];
  \legend{\Large{Target $\rvp_2$}, , ,\Large{$\rvp_2$ of generated images}}
 \end{axis}
 \end{tikzpicture}} 
 \caption{{$2^{nd}$ moment curves}}
 \label{fig:p2_curve}
\end{subfigure}
\begin{subfigure}[b]{0.295\textwidth}
          \centering
         \resizebox{\linewidth}{!}{
\begin{tikzpicture}
 \begin{axis}[ymin=0,axis background/.style={fill=seabornback, fill opacity=1},
grid style={line width=.1pt, draw=white},
    major grid style={line width=.1pt,draw=white},
    minor tick num=1,
     grid=both,
     xlabel=\Large{Volume fraction ($p_1$)},
     xmin=0.415,
     xmax=0.455,
     ylabel=\Large{Sample density},
     legend cell align=left,
     legend style={at={(1,1.25)}},ylabel style={at={(2ex,0.5)}}]
     \draw [ultra thick, color=ube]({axis cs:0.436,0}|-{rel axis cs:0,0}) -- ({axis cs:0.436,0}|-{rel axis cs:0,1});
\addlegendimage{color=ube,thick}
\addlegendentry{\Large{Target $p_1$}}
\addplot[thick,color=amber,fill, fill opacity=0.2,area legend] table[x=x, y=kde, col sep=comma]{data/hybrid_p1.csv}; \addlegendentry{\Large{$p_1$ of generated images}}
\end{axis}
\end{tikzpicture}}
 \caption{\small{Distribution of $1^{st}$ moment ($p_1$)}}
\end{subfigure}
\caption{\sl Comparisons of $1^{st}$ moment ($p_1$) distribution and $2^{nd}$ moment curves between the images generated by InvNet and the target invariances.}
     \label{fig:hybridgraph}
 \end{figure}

\renewcommand{\wa}{0.055}
\begin{figure}[t!]
	\begin{center}
		\setlength{\tabcolsep}{1pt}
		\renewcommand{\arraystretch}{0.5}
		\begin{tabular}{ccccccccccccccccc}
			\raisebox{0.8em}{\small{(a)}} &
			\includegraphics[width=0.054\linewidth,cframe=ube 1pt]{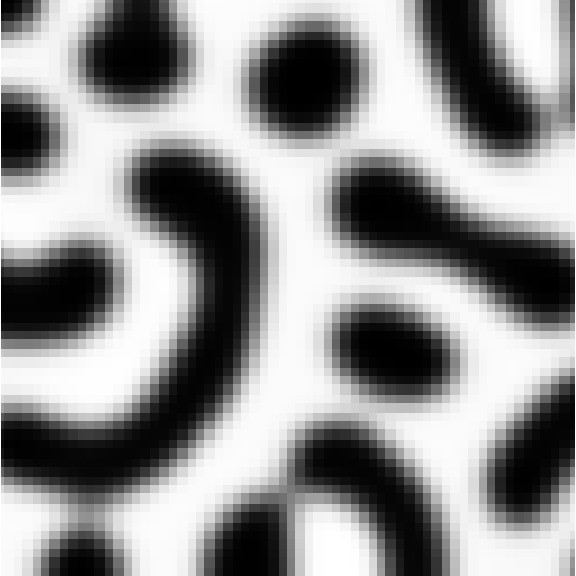} &
			\includegraphics[width=\wa\linewidth]{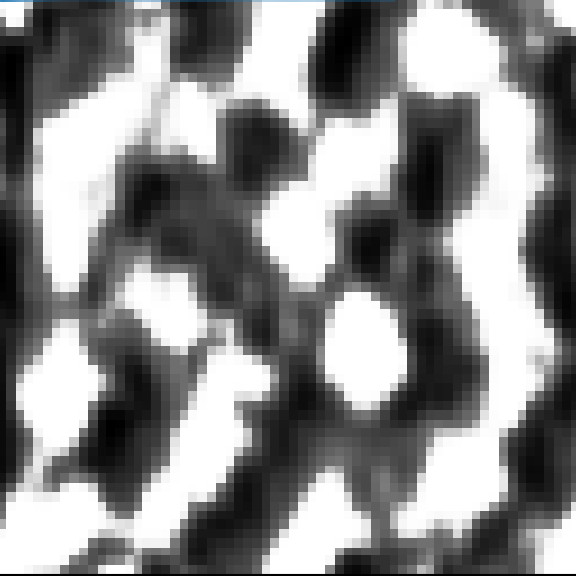} &
			\includegraphics[width=\wa\linewidth]{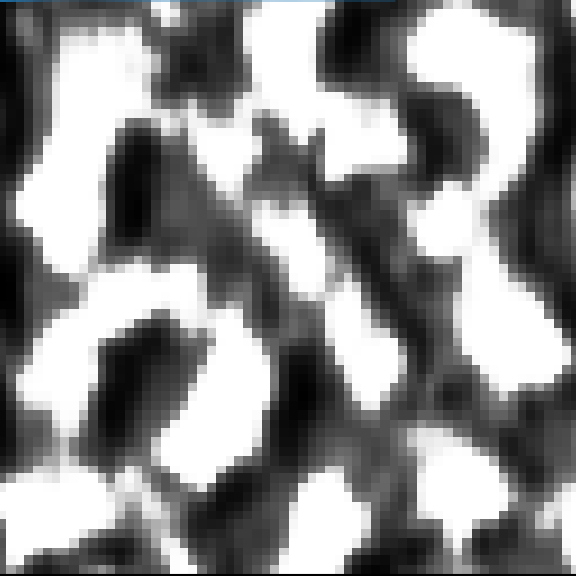} &
			\includegraphics[width=\wa\linewidth ]{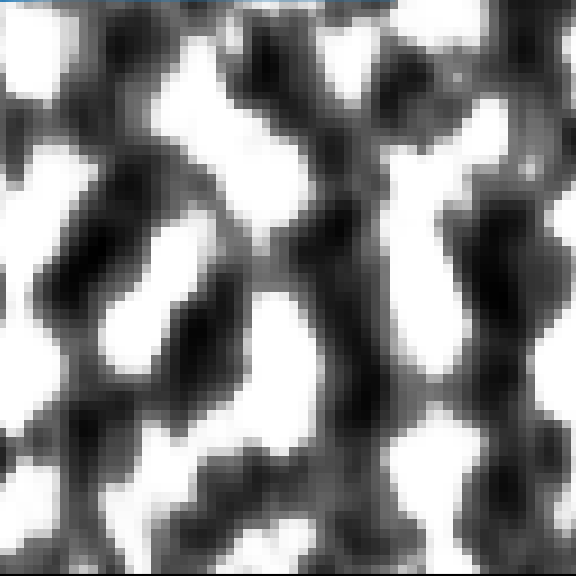} &
			\includegraphics[width=\wa\linewidth ]{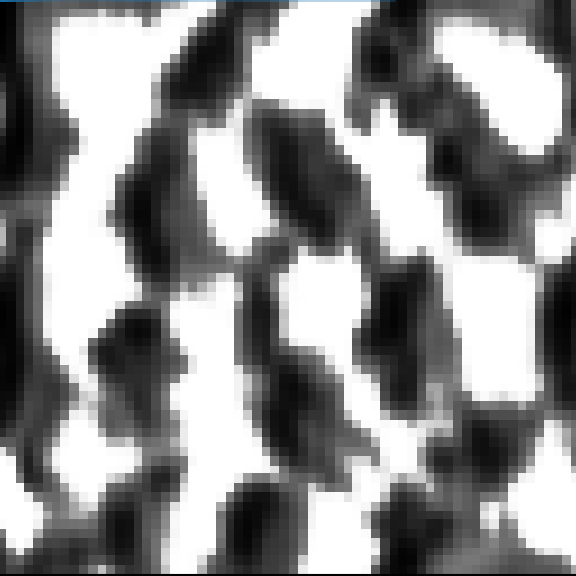} &
			\includegraphics[width=\wa\linewidth ]{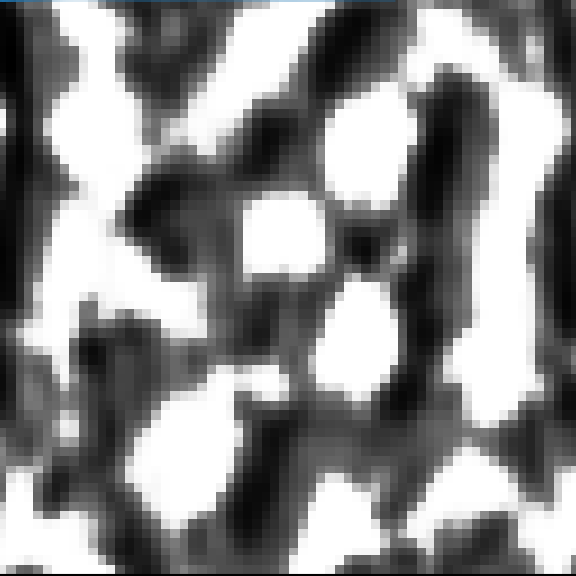} &
			\includegraphics[width=\wa\linewidth ]{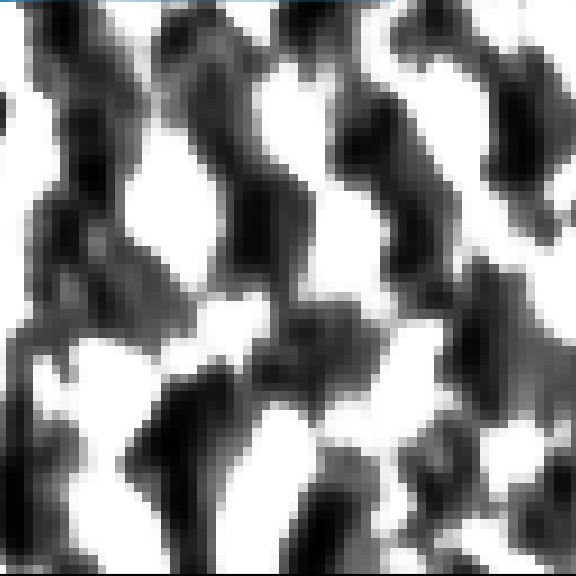} &
			\includegraphics[width=\wa\linewidth ]{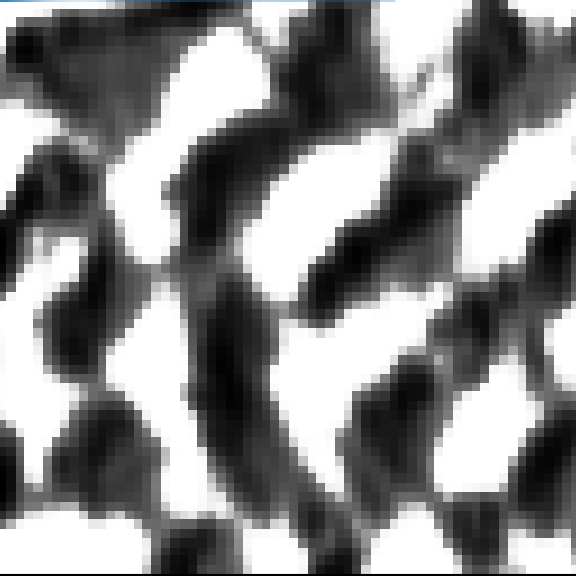} &
			\includegraphics[width=\wa\linewidth ]{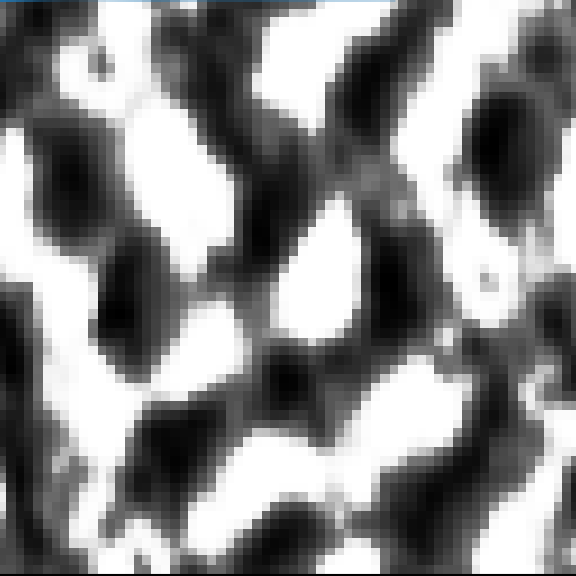} &
			\includegraphics[width=\wa\linewidth ]{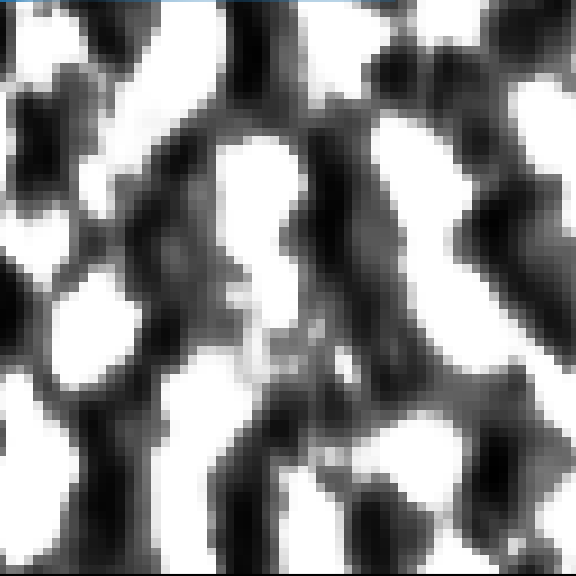} &
			\includegraphics[width=\wa\linewidth ]{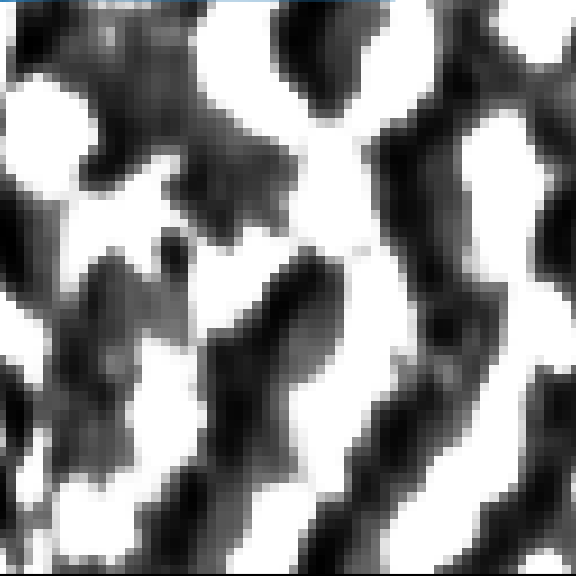} &
			\includegraphics[width=\wa\linewidth ]{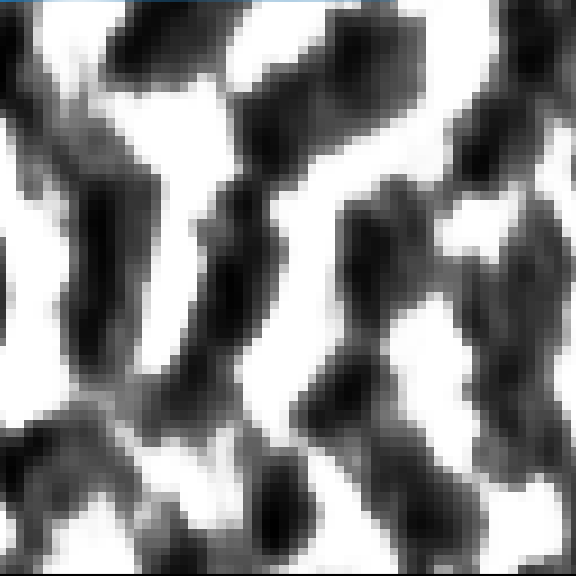} & 
			\includegraphics[width=\wa\linewidth ]{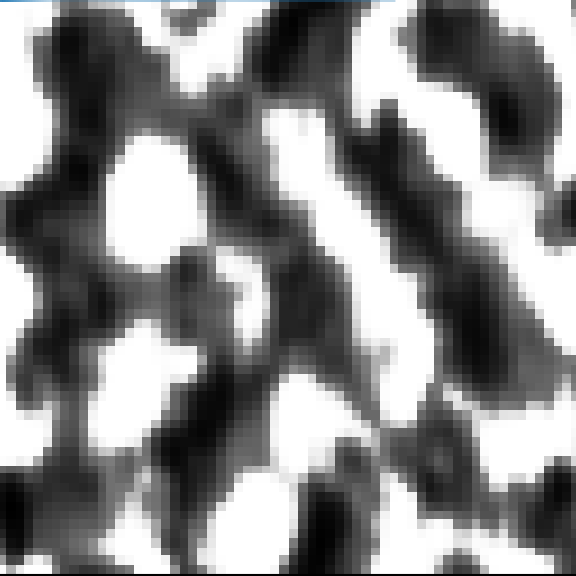} &
			\includegraphics[width=\wa\linewidth ]{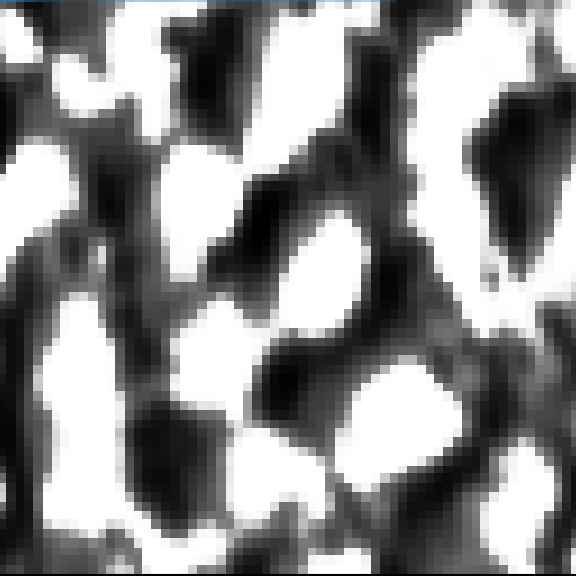} &
			\includegraphics[width=\wa\linewidth ]{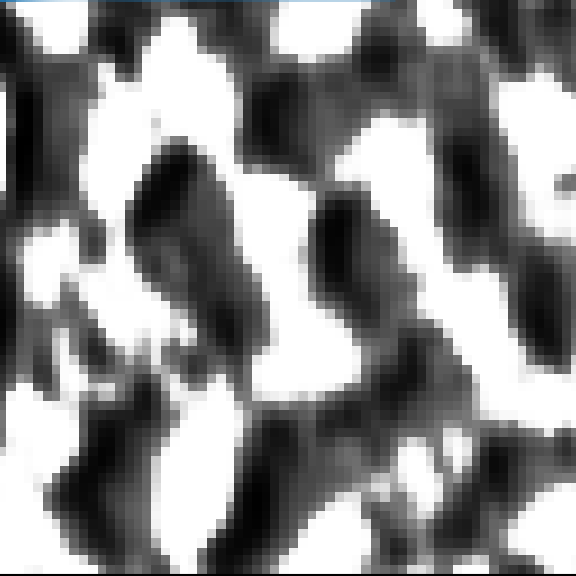} &	
			\includegraphics[width=\wa\linewidth ]{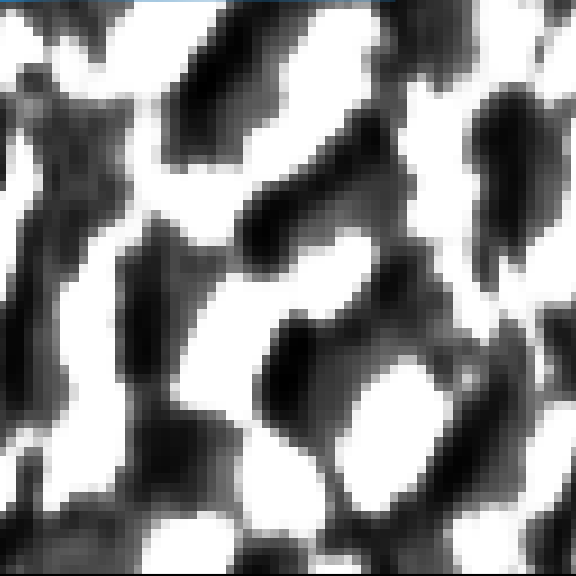} \\ 
			
			\raisebox{0.8em}{\small{(b)}} &
			\includegraphics[width=0.054\linewidth , cframe=ube 1pt]{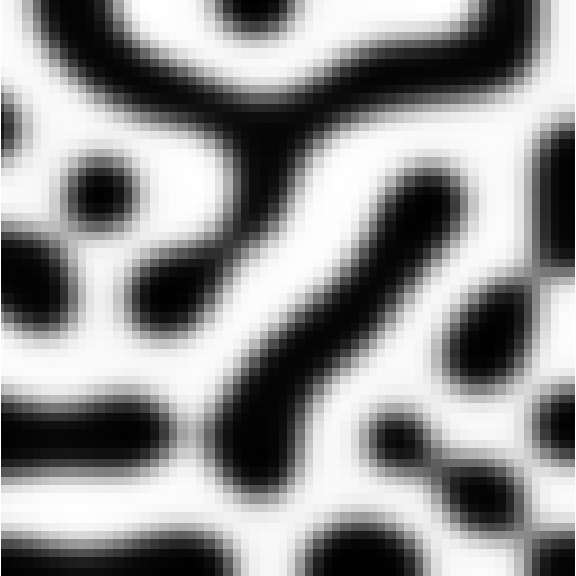} &
			\includegraphics[width=\wa\linewidth ]{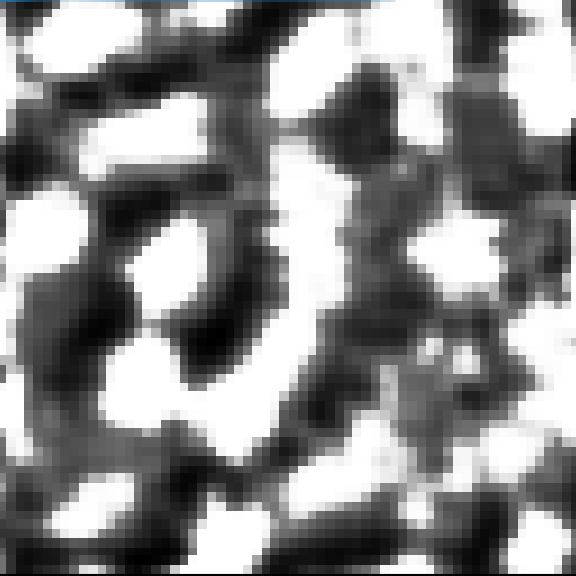} &
			\includegraphics[width=\wa\linewidth ]{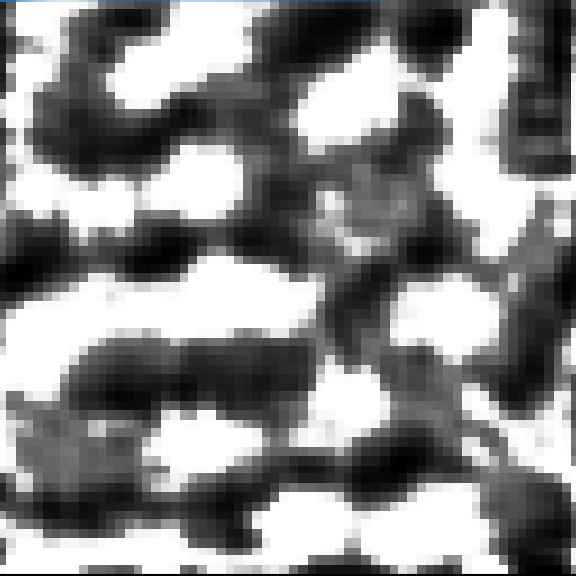} &
			\includegraphics[width=\wa\linewidth ]{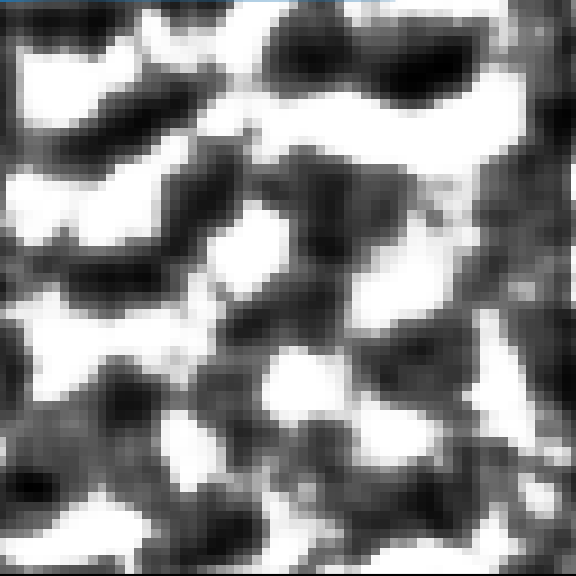} &
			\includegraphics[width=\wa\linewidth ]{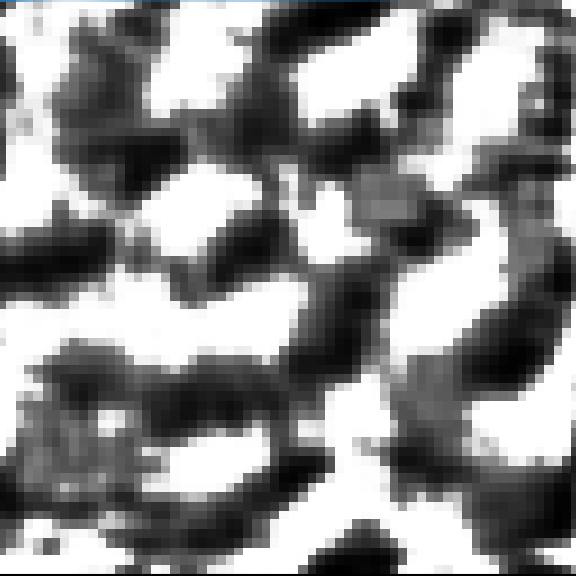} &
			\includegraphics[width=\wa\linewidth ]{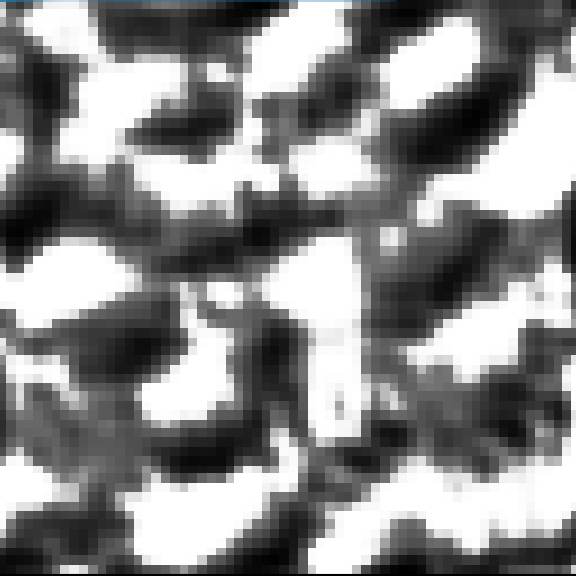} &
			\includegraphics[width=\wa\linewidth ]{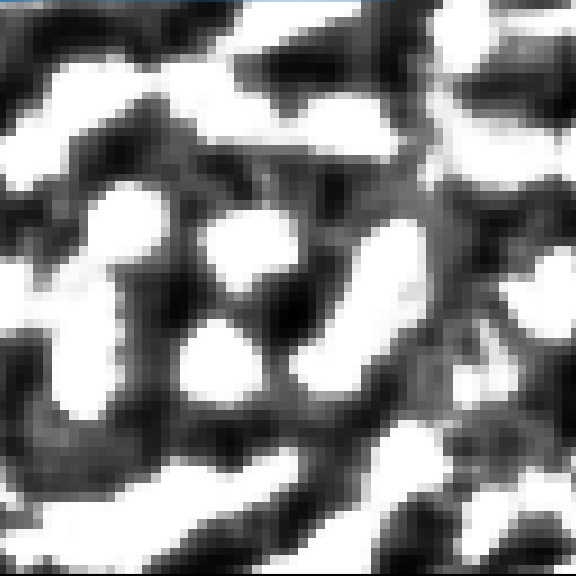} &
			\includegraphics[width=\wa\linewidth ]{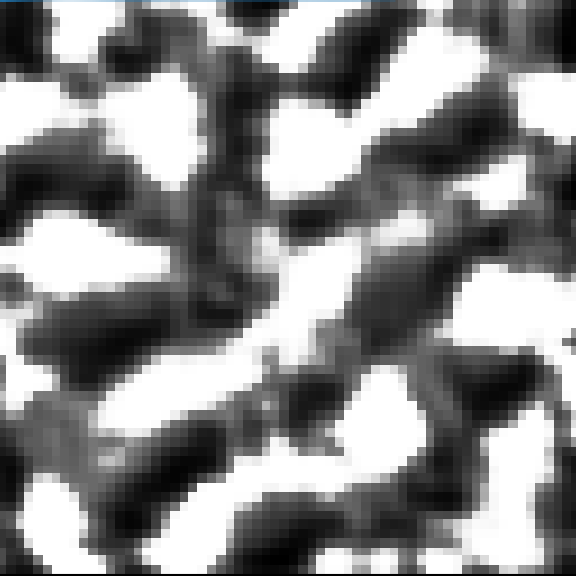} &
			\includegraphics[width=\wa\linewidth ]{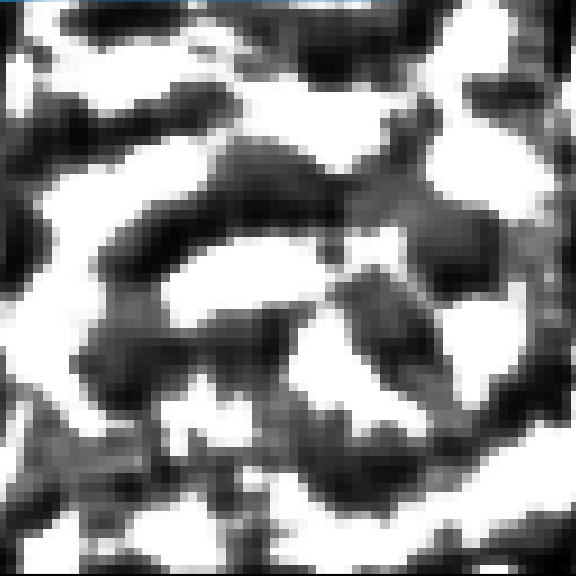} &
			\includegraphics[width=\wa\linewidth ]{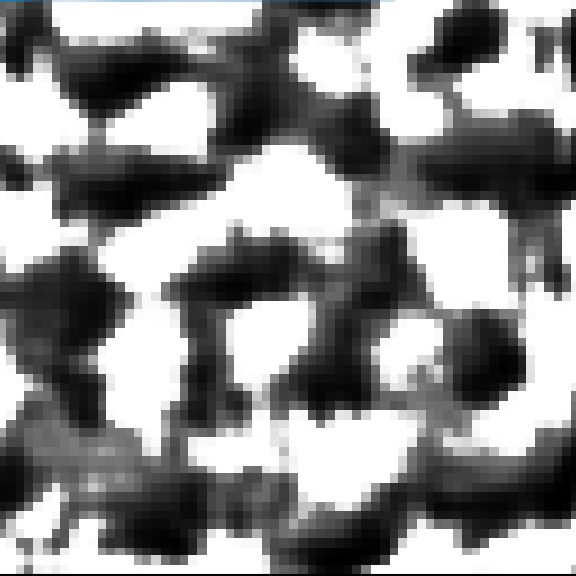} &
			\includegraphics[width=\wa\linewidth ]{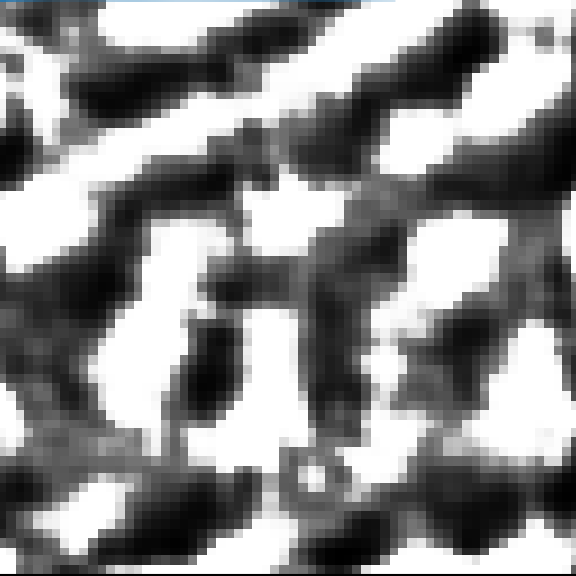} &
			\includegraphics[width=\wa\linewidth ]{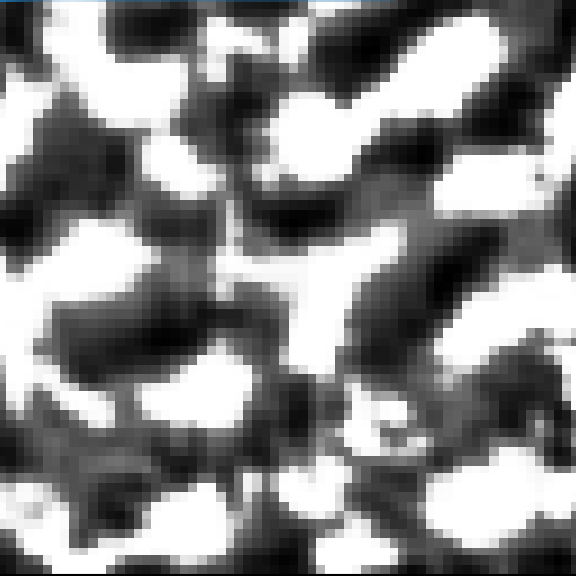} & 
			\includegraphics[width=\wa\linewidth ]{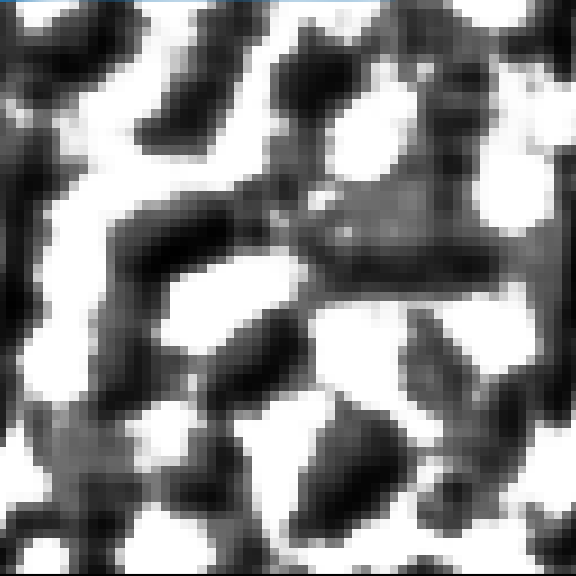} &
			\includegraphics[width=\wa\linewidth ]{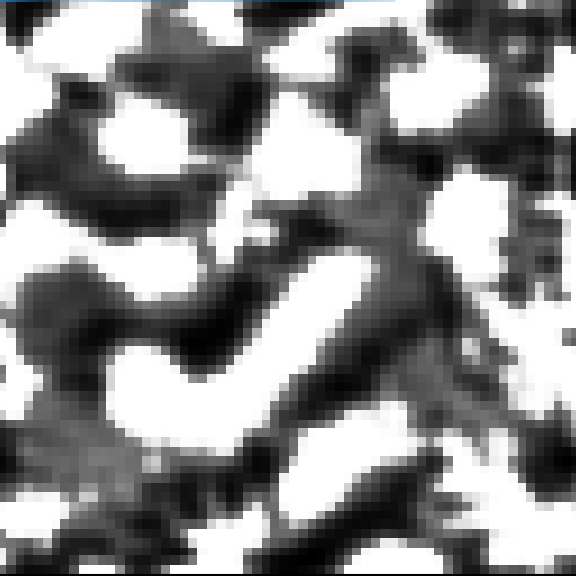} &
			\includegraphics[width=\wa\linewidth ]{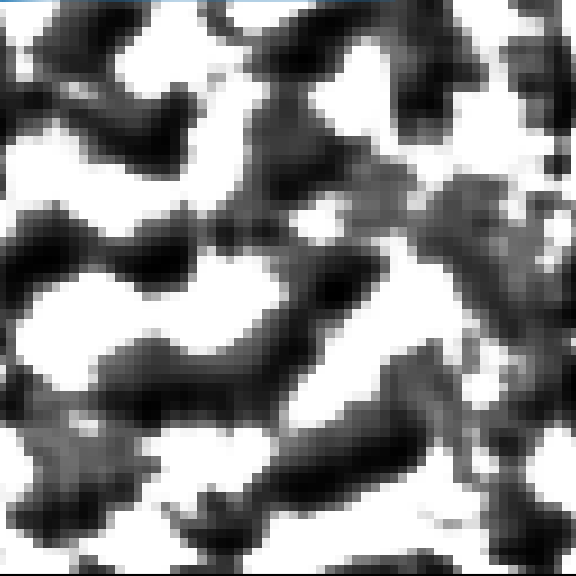} &	
			\includegraphics[width=\wa\linewidth ]{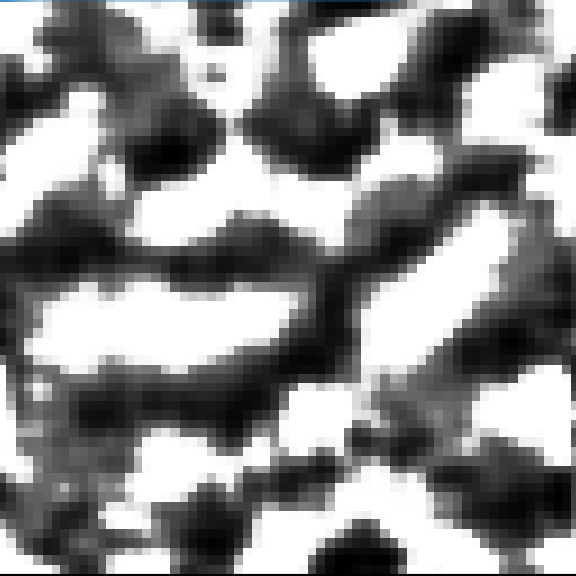} \\
			
			\raisebox{0.8em}{\small{(c)}} &
			\includegraphics[width=0.054\linewidth , cframe=ube 1pt]{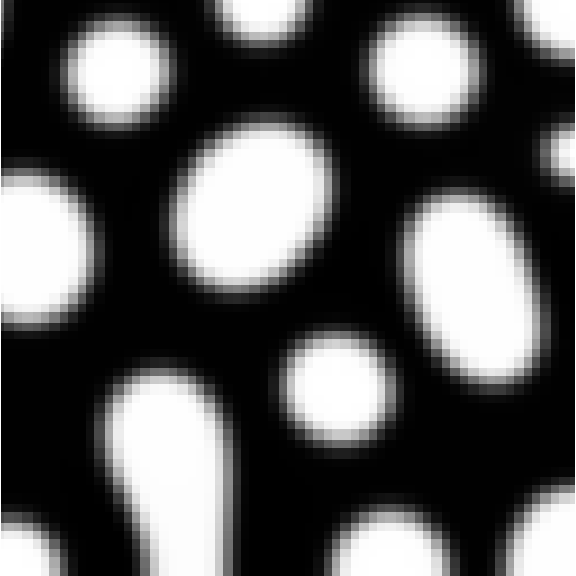} &
			\includegraphics[width=\wa\linewidth ]{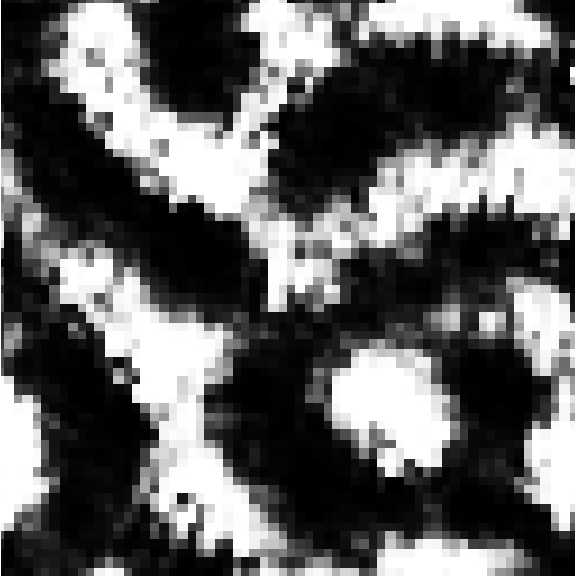} &
			\includegraphics[width=\wa\linewidth ]{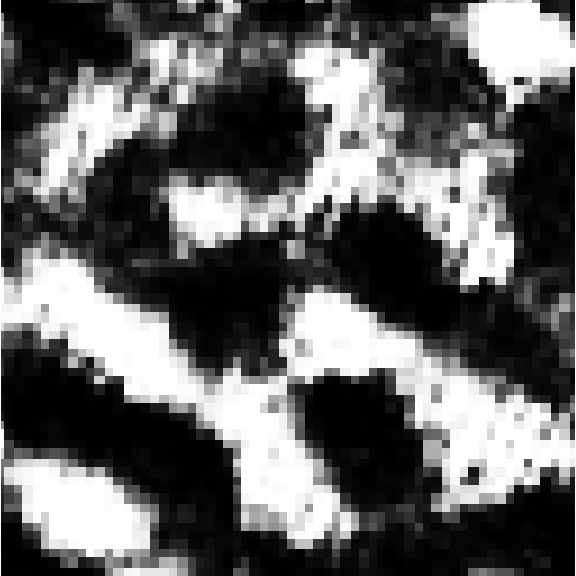} &
			\includegraphics[width=\wa\linewidth ]{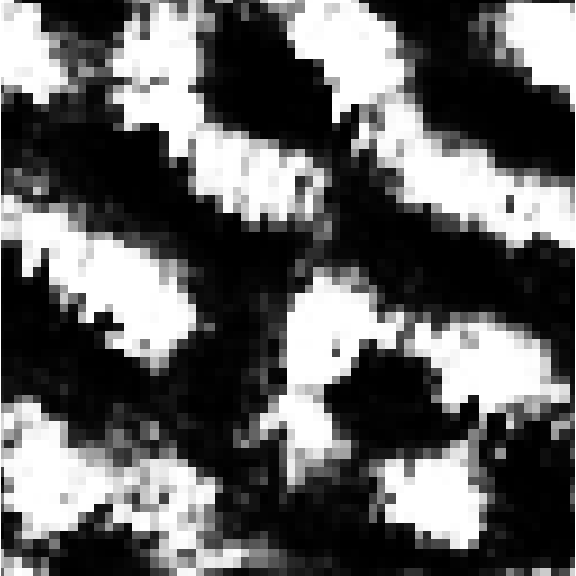} &
			\includegraphics[width=\wa\linewidth ]{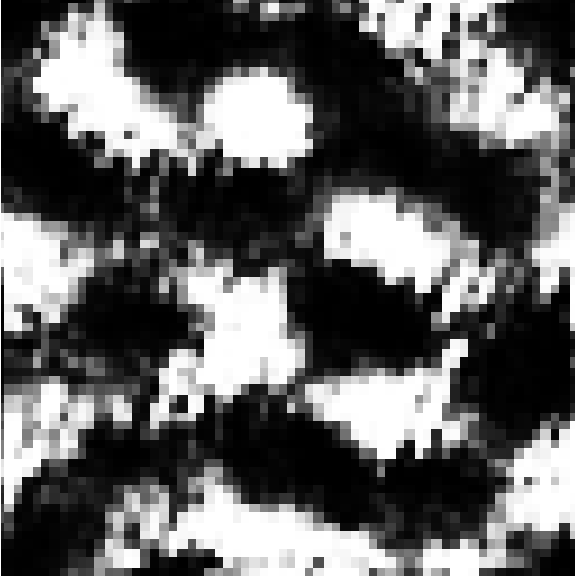} &
			\includegraphics[width=\wa\linewidth ]{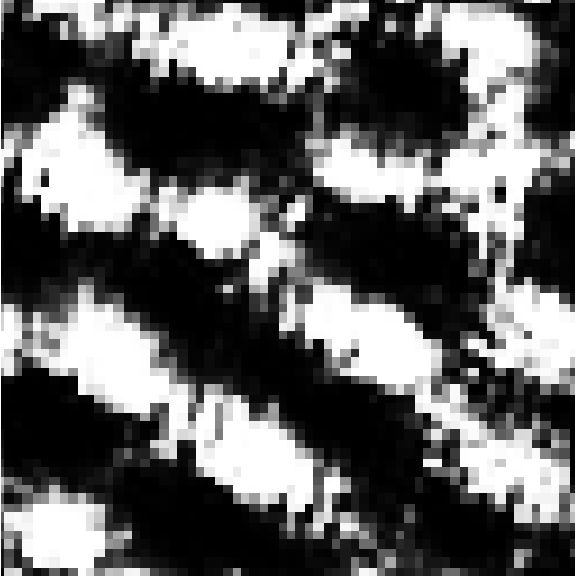} &
			\includegraphics[width=\wa\linewidth ]{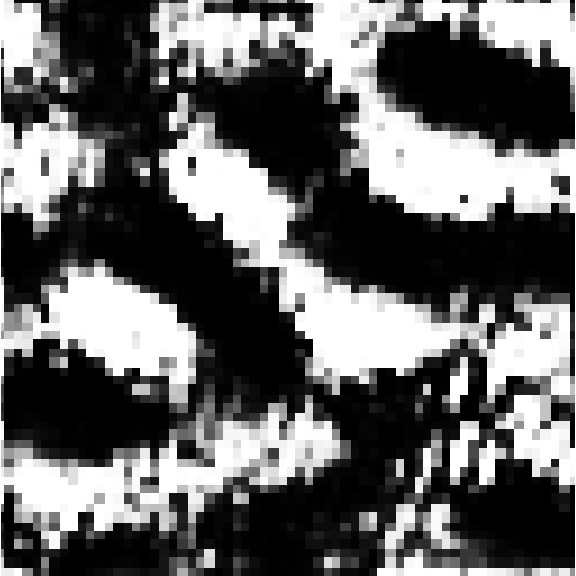} &
			\includegraphics[width=\wa\linewidth ]{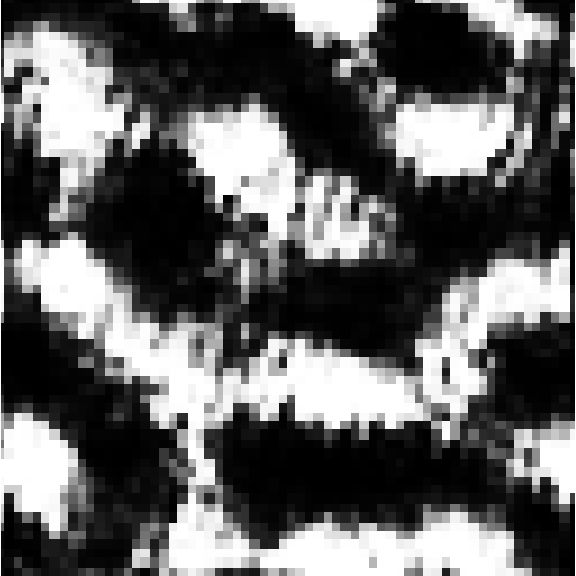} &
			\includegraphics[width=\wa\linewidth ]{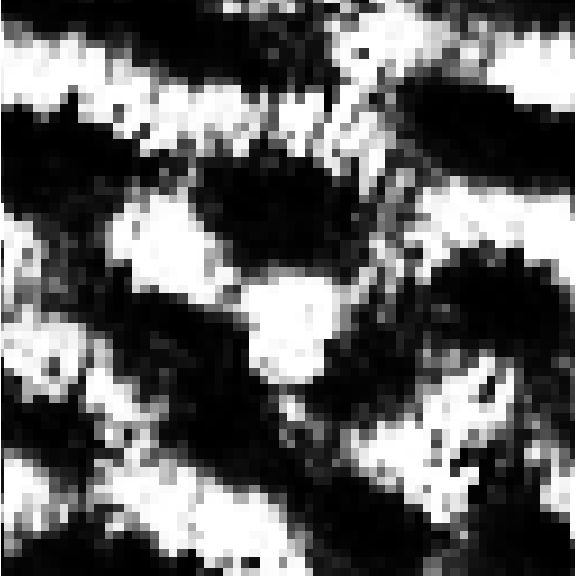} &
			\includegraphics[width=\wa\linewidth ]{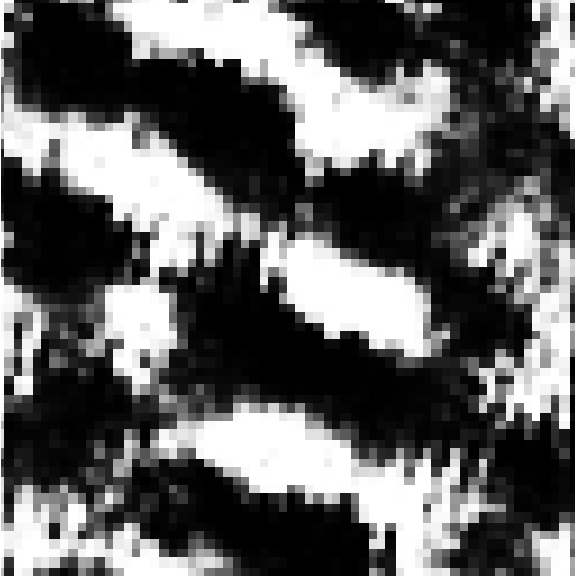} &
			\includegraphics[width=\wa\linewidth ]{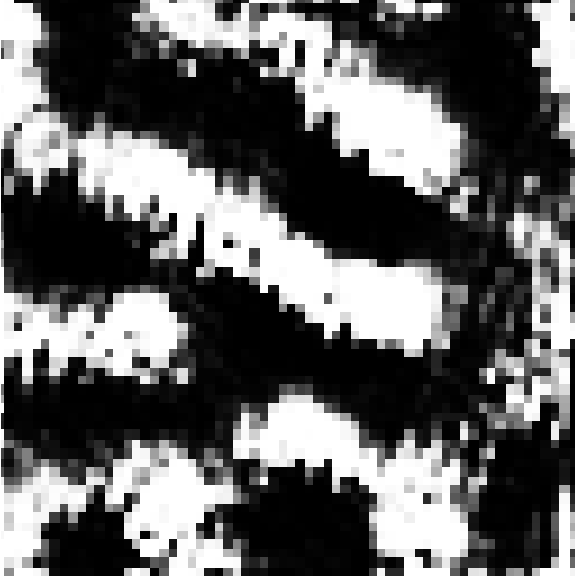} &
			\includegraphics[width=\wa\linewidth ]{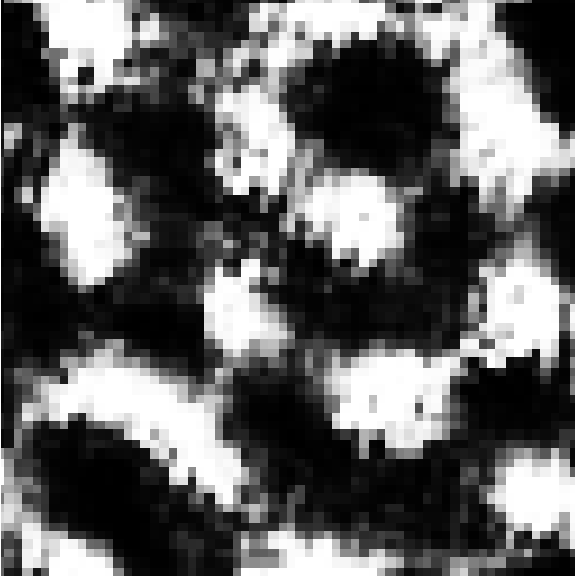} & 
			\includegraphics[width=\wa\linewidth ]{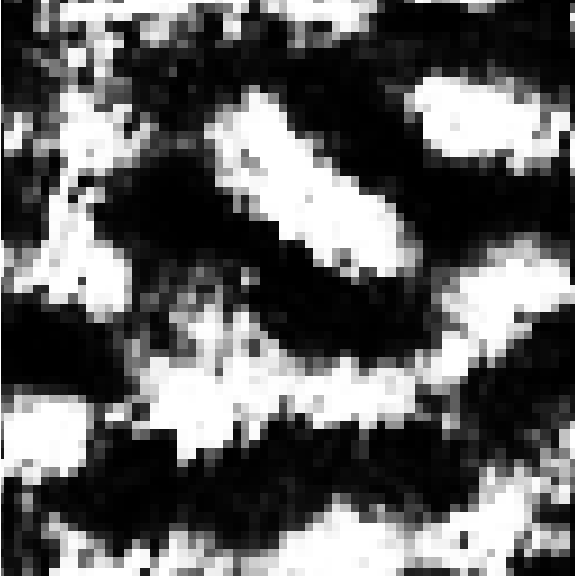} &
			\includegraphics[width=\wa\linewidth ]{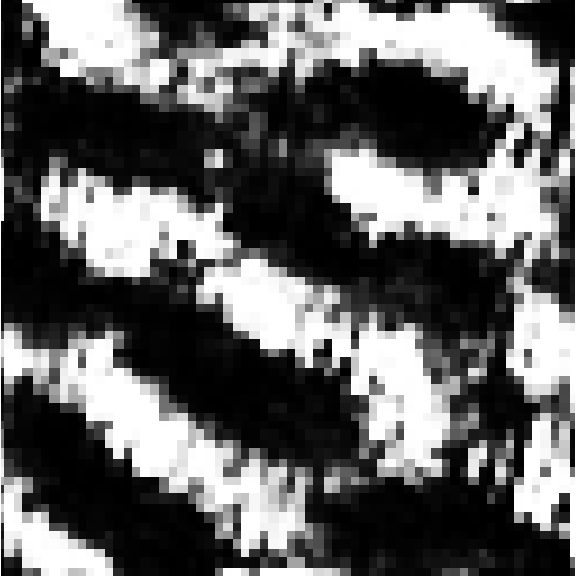} &
			\includegraphics[width=\wa\linewidth ]{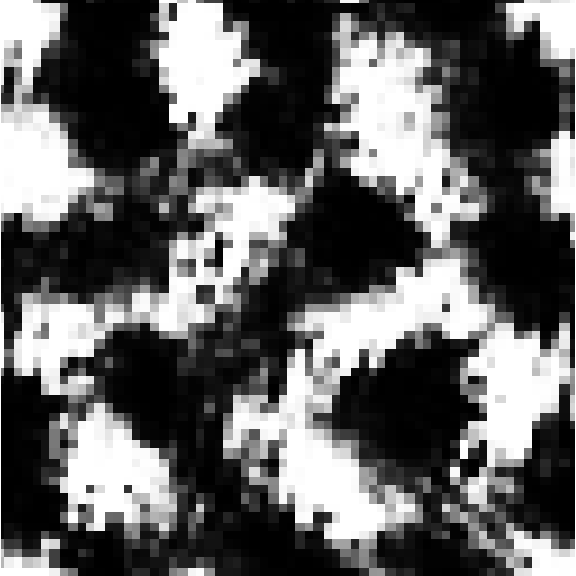} &	
			\includegraphics[width=\wa\linewidth ]{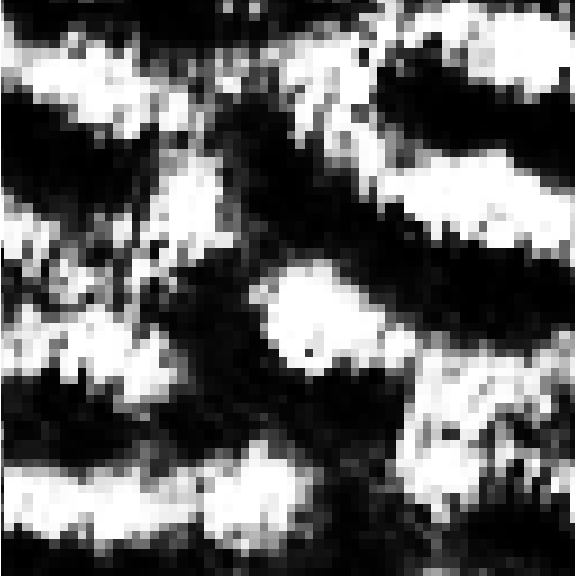} \\
		\end{tabular}
	\end{center}
	\caption{{\sl Images generated by InvNet trained only with invariance loss; with first image in each row being the real image used to calculate the target invariances. (a,b) are trained over entire $2^{nd}$ moment curve while model in (c) used only the initial portion of $2^{nd}$ moment curve.}}
	\label{fig:invar1}
\end{figure}

\pgfkeys{/pgf/number format/.cd,1000 sep={\,}}
\usepgfplotslibrary{fillbetween}
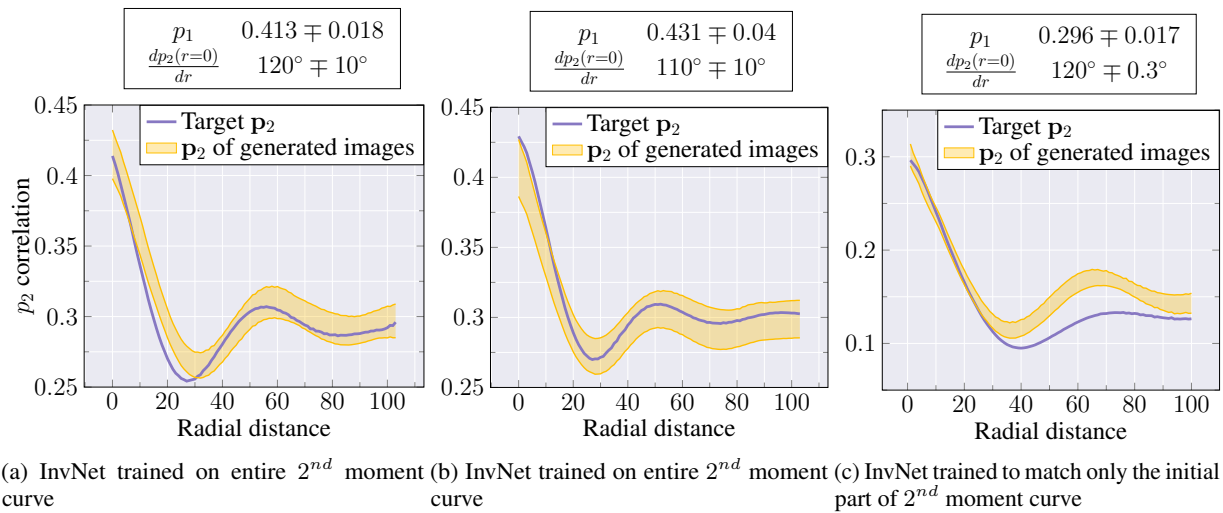
\begin{figure}[t!]
	\begin{subfigure}[b]{0.34\textwidth}
		\centering
		\resizebox{\linewidth}{!}{
			\begin{tikzpicture}
			\tikzstyle{every node}=[font=\Large] 
			\begin{axis}[axis background/.style={fill=seabornback, fill opacity=1},
			ymin=0.25,
			ymax=0.45,
			ylabel=$p_2$ correlation,
			grid style={line width=.1pt, draw=white},
			major grid style={line width=.1pt,draw=white},
			minor tick num=1,
			grid=both,
			xlabel= Radial distance,
			legend cell align=left,
			legend style={at={(1,1)}}]
			\addplot[ultra thick,color=ube] table[x=index, y=real, col sep=comma]{data/inv_p2_1_corrected.csv};
			\addplot[thick,name path=max,color=amber] table[x=index, y=max, col sep=comma]{data/inv_p2_1_corrected.csv};
			\addplot[thick,name path = min, color=amber] table[x=index, y=min, col sep=comma]{data/inv_p2_1_corrected.csv};
			\addplot[color=amber, fill opacity=0.3] fill between[
			of = max and min];
			\legend{Target $\rvp_2$, , ,$\rvp_2$ of generated images}
			\end{axis} 
			\draw[shorten >=1mm,shorten <=1mm] (rel axis cs:0.20,1.20) node[right,draw,align=left]{
				\renewcommand{\arraystretch}{1.2}
				\begin{tabular}{cc}
				$p_1$ & $0.413 \mp 0.018$ \\
				$\frac{dp_2(r=0)}{dr}$  & $120\degree \mp 10\degree$ \\
				\end{tabular}};
			\end{tikzpicture}} 
		\caption{\footnotesize{InvNet trained on entire $2^{nd}$ moment curve}}
	\end{subfigure}
	\begin{subfigure}[b]{0.32\textwidth}
		\centering
		\resizebox{\linewidth}{!}{
			\begin{tikzpicture}
			\tikzstyle{every node}=[font=\Large]
			\begin{axis}[axis background/.style={fill=seabornback, fill opacity=1},
			ymin=0.25,
			ymax=0.45,
			xlabel= Radial distance,
			grid style={line width=.1pt, draw=white},
			major grid style={line width=.1pt,draw=white},
			minor tick num=1,
			grid=both,
			legend cell align=left,
			legend style={at={(1,1)}}]
			\addplot[ultra thick,color=ube] table[x=index, y=real, col sep=comma]{data/inv_p2_2_corrected.csv};
			\addplot[thick,name path=max,color=amber] table[x=index, y=max, col sep=comma]{data/inv_p2_2_corrected.csv};
			\addplot[thick,name path = min, color=amber] table[x=index, y=min, col sep=comma]{data/inv_p2_2_corrected.csv};
			\addplot[color=amber, fill opacity=0.3] fill between[
			of = max and min];
			\legend{Target $\rvp_2$, , ,$\rvp_2$ of generated images}
			\end{axis}
			\draw[shorten >=1mm,shorten <=1mm] (rel axis cs:0.20,1.20) node[right,draw,align=left]{
				\renewcommand{\arraystretch}{1.2}
				\begin{tabular}{cc}
				$p_1$ & $0.431 \mp 0.04$ \\
				$\frac{dp_2(r=0)}{dr}$  & $110\degree \mp 10\degree$ \\
				\end{tabular}};
			\end{tikzpicture}}
		\caption{\footnotesize{InvNet trained on entire $2^{nd}$ moment curve}}
	\end{subfigure}
	\begin{subfigure}[b]{0.31\textwidth}
		\centering
		\resizebox{\linewidth}{!}{
			\begin{tikzpicture}
			\tikzstyle{every node}=[font=\Large]
			\begin{axis}[axis background/.style={fill=seabornback, fill opacity=1},
			ymin=0.05,
			ymax=0.35,
			grid style={line width=.1pt, draw=white},
			major grid style={line width=.1pt,draw=white},
			minor tick num=1,
			grid=both,
			xlabel=Radial distance,
			legend cell align=left,
			legend style={at={(1,1)}}]
			\addplot[ultra thick,color=ube] table[x=index, y=real, col sep=comma]{data/inv_p2_4.csv};
			\addplot[thick,name path=max,color=amber] table[x=index, y=max, col sep=comma]{data/inv_p2_4.csv};
			\addplot[thick,name path = min, color=amber] table[x=index, y=min, col sep=comma]{data/inv_p2_4.csv};
			\addplot[color=amber, fill opacity=0.3,area legend] fill between[
			of = max and min];
			\legend{Target $\rvp_2$, , ,$\rvp_2$ of generated images}
			\end{axis}
			\draw[shorten >=1mm,shorten <=1mm] (rel axis cs:0.20,1.20) node[right,draw,align=left]{
				\renewcommand{\arraystretch}{1.2}
				\begin{tabular}{cc}
				$p_1$ & $0.296 \mp 0.017$ \\
				$\frac{dp_2(r=0)}{dr}$  & $120\degree \mp 0.3\degree$ \\
				\end{tabular}};
			\end{tikzpicture}}
		\caption{\footnotesize{InvNet trained to match only the initial part of $2^{nd}$ moment curve}}
	\end{subfigure}
	\caption{\sl Comparisons of $2^{nd}$ moment curves between the images generated by InvNet trained only with invariance loss and the target invariances.}
	\label{fig:invar2a}
\end{figure}

\end{document}